\documentclass{nature}

\usepackage{graphicx}
\usepackage{xcolor}
\usepackage{url}
\usepackage[normalem]{ulem}
\usepackage{comment}
\usepackage{caption}
\usepackage{subcaption}
\usepackage{amsfonts}
\usepackage{amsmath}
\usepackage{placeins}
\usepackage{verbatim}
\usepackage{calc}
\usepackage{hyperref}
\usepackage[style=nature]{biblatex}
\usepackage{lineno}
\makeatletter

\csnumgdef{blx@entrycount}{0}
\AtEveryBibitem{%
  \csnumgdef{blx@entrycount}{\csuse{blx@entrycount}+1}}

\appto{\newrefsegment}{%
  \csnumgdef{blx@entrycount@\the\c@refsegment}{\csuse{blx@entrycount}+1}}

\defbibcheck{onlynew}{%
  \ifnumless{\thefield{labelnumber}}{\csuse{blx@entrycount@\the\c@refsegment}}
    {\skipentry}
    {}}

\makeatother

\newcommand{\print}[1]{\csname #1\endcsname: \csname #1@bar\endcsname}

\renewcommand*{\cite}[1]{\supercite{#1}}

\usepackage{setspace}

\iftrue
\newcommand{\comments}[1]{#1}

\else
\newcommand{\comments}[1]{}

\fi

\newcommand{\citefull}[1]{\citeauthor{#1}~(\citeyear{#1})\cite{#1}}

\definecolor{maroon}{rgb}{.5,0,0}
\definecolor{deepForestGreen}{rgb}{.2,.8,.04}
\definecolor{magenta}{rgb}{0.7,0,1}

\usepackage[symbol]{footmisc}

\title{First return, then explore}

\author{Adrien Ecoffet$^{*1}$, Joost Huizinga$^{*1}$, Joel Lehman$^1$, Kenneth O. Stanley$^1$ \& Jeff Clune$^{1,2}$}

\begin{document}

\maketitle

\begin{affiliations}
 \item Uber AI, San Francisco, CA, USA
 \item OpenAI, San Francisco, CA, USA (work done at Uber AI)

$^*$ These authors contributed equally to this work
\end{affiliations}

\vspace{5mm}
\noindent
\textbf{Authors' note}: This is the pre-print version of this work. You most likely want the updated, published version, which can be found as an unformatted PDF at \href{https://adrien.ecoffet.com/files/go-explore-nature.pdf}{\texttt{https://adrien.ecoffet.com/files/go-explore-nature.pdf}} or as a formatted web-only document at \href{https://tinyurl.com/Go-Explore-Nature}{\texttt{https://tinyurl.com/Go-Explore-Nature}}.

\noindent Please cite as:\\
Ecoffet, A., Huizinga, J., Lehman, J., Stanley, K.O. and Clune, J. First return, then explore. \emph{Nature} \textbf{590,} 580–586 (2021). \href{https://doi.org/10.1038/s41586-020-03157-9}{https://doi.org/10.1038/s41586-020-03157-9}
\vspace{15mm}

\ifarxiv
\else
\linenumbers
\fi

\newrefsegment
\begin{abstract}
The promise of reinforcement learning is
to solve complex sequential decision problems autonomously by specifying a high-level reward function only.
However, 
reinforcement learning algorithms struggle when, as is often the case, simple and intuitive rewards provide sparse\cite{bellemare2016unifying} and deceptive\cite{lehman} feedback.
Avoiding these pitfalls requires thoroughly exploring the environment, but despite substantial investments by the community, creating algorithms that can do so remains one of the central challenges of the field.
We hypothesise that the main impediment to effective exploration originates from algorithms forgetting how to reach previously visited states (``detachment'') and from failing to first return to a state before exploring from it (``derailment''). 
We introduce Go-Explore, a family of algorithms that addresses these two challenges directly through the simple principles of
explicitly remembering promising states and first returning to such states before intentionally exploring.
Go-Explore solves all heretofore unsolved Atari games 
(meaning those for which algorithms could not previously outperform humans when evaluated following current community standards for Atari\cite{Machado2018RevisitingTA})
and surpasses the state of the art on all hard-exploration games\cite{bellemare2016unifying}, with orders of magnitude improvements on the grand challenges \emph{Montezuma's Revenge} and \emph{Pitfall}.
We also demonstrate the practical potential of Go-Explore on a challenging and extremely sparse-reward pick-and-place robotics task. Additionally, we show that adding a goal-conditioned policy can further improve Go-Explore's exploration efficiency and enable it to handle stochasticity throughout training.
The striking contrast between the substantial performance gains from Go-Explore and the simplicity of its mechanisms suggests that remembering promising states, returning to them, and exploring from them is a powerful and general approach to exploration, an insight that may prove critical to the  creation of truly intelligent learning agents.

\end{abstract}

Recent years have yielded impressive achievements in Reinforcement Learning (RL), including world-champion level performance in Go\cite{Silver2017MasteringTG}, Starcraft II\cite{vinyals2019grandmaster}, and Dota II\cite{berner2019dota}, as well as autonomous learning of robotic skills such as running, jumping, and grasping\cite{merel2018hierarchical, andrychowicz2020learning}.
Many of these successes were enabled by reward functions that are 
carefully designed to be highly informative.
However, for many practical problems, defining a good reward function is non-trivial;
to guide a household robot to a coffee machine, one might provide a reward only when the machine is successfully
reached, but doing so makes the reward \emph{sparse}, as it may require many correct actions for the coffee machine to be reached in the first place.
One might instead attempt to define a denser reward, such as the Euclidean distance towards the machine, but such a reward function can be \emph{deceptive}; naively following the reward function may lead the robot into a dead end that does not lead to the machine.
Furthermore, attempts to define a dense reward function are a common cause of \emph{reward hacking}, in which the agent finds a way to maximise the reward function by producing wholly unintended (and potentially unsafe) behaviour\cite{Lehman2018TheSC,Amodei2016ConcretePI}, such as knocking over furniture or driving through a wall to reach the machine.

These challenges motivate designing RL algorithms that better handle sparsity and deception.
A key observation is that sufficient \emph{exploration} of the state space enables discovering sparse rewards as well as avoiding deceptive local optima\cite{lehman2011abandoning, conti2018improving}. 
We argue that two major issues have hindered the ability of previous algorithms to explore: \emph{detachment}, in which the algorithm loses track of interesting areas to explore from, 
and \emph{derailment}, in which the exploratory mechanisms of the algorithm prevent it from returning to previously visited states, preventing exploration directly and/or forcing practitioners to make exploratory mechanisms so minimal that effective exploration does not occur (Supplementary Information).
We present Go-Explore, a family of algorithms designed to explicitly avoid detachment and derailment.
We demonstrate how the Go-Explore paradigm allows the creation of algorithms that thoroughly explore environments. 
Go-Explore solves all previously unsolved Atari games\footnote{Concurrent work\cite{badia2020agent57} has similarly reached this milestone on all Atari games, but was evaluated under easier, mostly deterministic conditions that do not meet community-defined standards\cite{Machado2018RevisitingTA} for evaluation on Atari (Methods ``\nameref{sec:atari_sota}''). Also, even if these differences in evaluation are ignored, Go-Explore accomplished the same feat at effectively the same time, but via a totally different method, which is useful for the scientific community. Go-Explore additionally obtains higher scores on most of the games that we tested (Supplementary Information).}, where solved is defined as surpassing human performance, which has been posited as a major milestone in previous work\cite{mnih:nature15, aytar2018playing, badia2020agent57}.
It also surpasses the state of the art on all hard-exploration Atari games (except in one case where the maximum score is reached by Go-Explore and previous algorithms).
Additionally, we demonstrate that it can solve a practical simulated robotics problem with an extremely sparse reward.
Finally, we show that its performance can be greatly increased by incorporating minimal domain knowledge and examine how harnessing a policy during exploration can improve exploration efficiency, highlighting the versatility of the Go-Explore family.

\section*{The Go-Explore family of algorithms}
\label{sec:goexplore_family}

\begin{figure}[tbp!]
    \centering
    \includegraphics[width=0.7\linewidth]{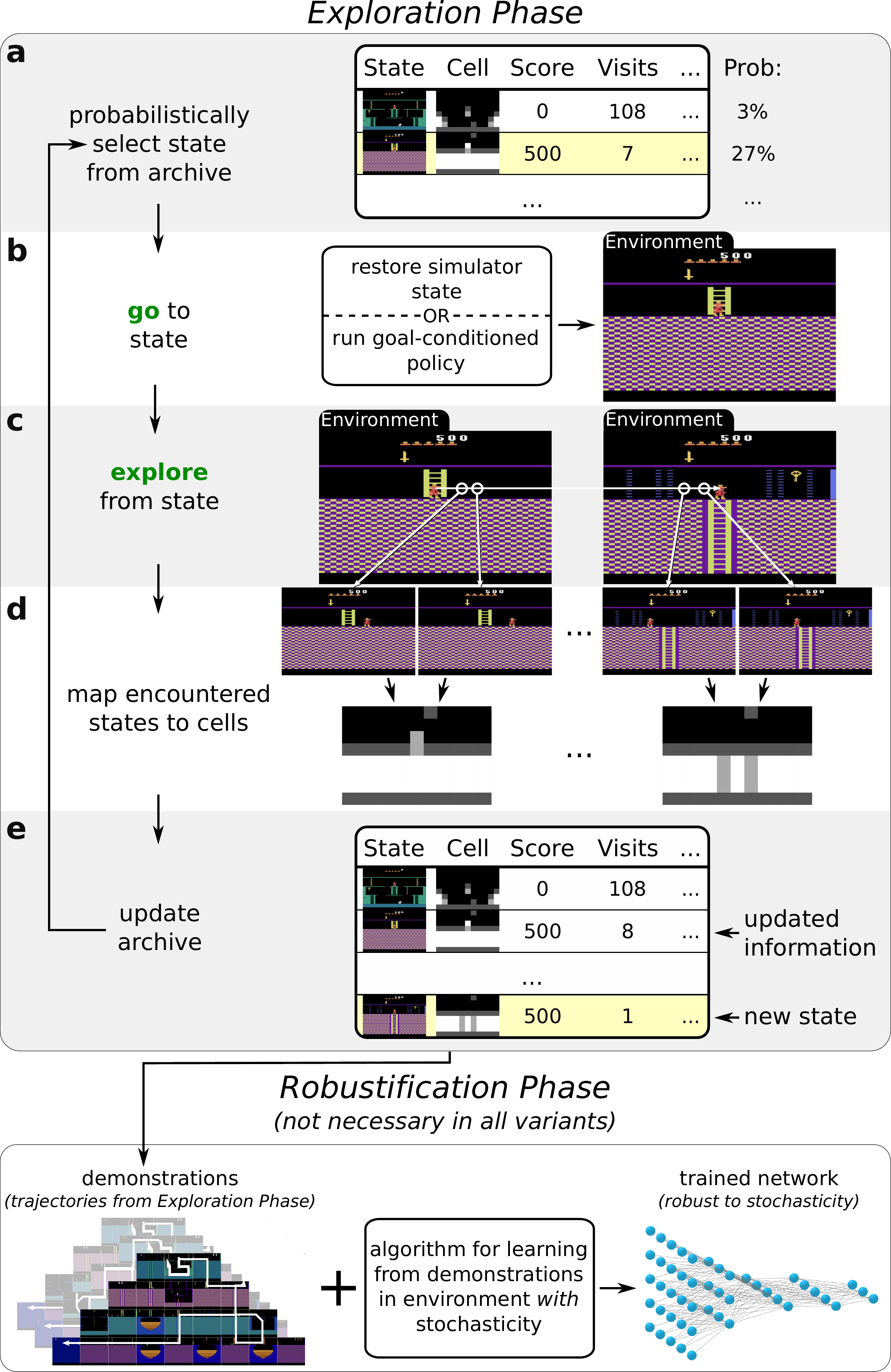}
    \caption{\textbf{Overview of Go-Explore.} \textbf{(a)} Probabilistically select a state from the archive, guided by heuristics that prefer states associated with promising cells. \textbf{(b)} Return to the selected state, such as by restoring simulator state or by running a goal-conditioned policy. \textbf{(c)} Explore from that state by taking random actions or sampling from a policy. \textbf{(d)} Map every state encountered during returning and exploring to a low-dimensional \emph{cell} representation. \textbf{(e)} Add states that map to new cells to the archive and update other archive entries.}
    \label{fig:concept_sketch}
\end{figure}

To avoid detachment, Go-Explore builds an \emph{archive} of the different states it has visited in the environment, thus ensuring that states cannot be forgotten. 
Starting from an archive containing the initial state, it builds this archive iteratively: first, it probabilistically selects a state to return to from the archive (Fig.~\ref{fig:concept_sketch}a), goes back (i.e. returns) to that state (the ``go'' step; Fig.~\ref{fig:concept_sketch}b), then explores from that state (the ``explore'' step; Fig.~\ref{fig:concept_sketch}c) and updates the archive with all novel states encountered (Fig.~\ref{fig:concept_sketch}e).
 Previous RL algorithms do not separate returning from exploring, and instead mix in exploration throughout an episode, usually by adding random actions a fraction of the time\cite{mnih:nature15,sutton1998reinforcement} or by sampling from a stochastic policy\cite{mnih2016asynchronous,Schulman2017ProximalPO}.
 By first returning before exploring, Go-Explore avoids derailment by minimizing exploration in the return policy (thus minimizing failure to return) after which it can switch to a purely exploratory policy. 

In practice, non-trivial environments have too many states to store explicitly.
To resolve this issue, similar states are grouped into \emph{cells}, and states are only considered novel if they are in a cell that does not yet exist in the archive (Fig.~\ref{fig:concept_sketch}d).
To maximise performance, if a state maps to an already known cell, but is associated with a better trajectory (e.g.\ higher performing or shorter, see Methods), that state and its associated trajectory will replace the state and trajectory currently associated with that cell. 
This behaviour is reminiscent of the MAP-Elites algorithm\cite{mouret2015illuminating}, and quality diversity algorithms more broadly\cite{lehman:gecco11,pugh:frontiers16}, although it is novel because it applies such ideas to reach new locations within a state space, instead of to discover new styles of behaviour as MAP-Elites typically does. 
Additionally, Go-Explore differs from MAP-Elites in that exploration is done by extending trajectories, rather than by modifying a policy and rerunning it.

While a policy can return the agent to a state (as we demonstrate at the end of this work), Go-Explore provides a unique opportunity to leverage the availability of \emph{simulators} in RL tasks; 
due to the large number of training trials current RL algorithms require, as well as the safety concerns that arise when running RL directly in the real world, simulators have played a key role in training the most compelling applications of RL\cite{cully2015robots,andrychowicz2020learning,peng2018sim,tan2018sim}, and will likely continue to be harnessed for the foreseeable future.
The inner state of any simulator can be saved and restored at a later time, making it possible to instantly return to a previously seen state.
However, returning without a policy 
also means that this exploration process, which we call the \emph{exploration phase}, does not produce a policy robust to the inherent stochasticity of the real world.
Instead, when returning without a policy, Go-Explore stores and maintains the highest-scoring trajectory (i.e.\ sequence of actions) to each cell.
After the exploration phase is complete, these trajectories make it possible to train a robust and high-performing policy by \emph{Learning from Demonstrations} (LfD)\cite{hester2017deep}, where the Go-Explore trajectories replace the usual human expert demonstrations.
Because it produces robust policies from brittle trajectories, we call this process the \emph{robustification phase} (Fig.~\ref{fig:concept_sketch} ``Robustification Phase'').

\section*{Learning to play Atari when returning without a policy}
\label{sec:deterministic}

The Atari benchmark suite\cite{bellemare2013arcade}, a prominent benchmark for RL algorithms\cite{mnih:nature15,horgan:apexdqn2018,espeholt:impala2018}, contains a diverse set of games with varying levels of reward sparsity and deceptiveness, making it an excellent platform for demonstrating the effectiveness of Go-Explore.
The following experiment highlights the benefit of a ``go'' step that restores the simulator state directly.
In this experiment, the ``explore'' step happens through random actions. The state-to-cell mapping consists of a parameter-based \emph{downscaling} of the current game frame, which results in similar-looking frames being aggregated into the same cell (Fig.~\ref{fig:concept_sketch}d). This mapping does not require any game-specific domain knowledge.
Good state-to-cell-mapping parameters result in a representation that strikes a balance between two extremes: lack of aggregation (e.g.\ one cell for every frame, which is computationally inefficient) and excessive aggregation (e.g.\ assigning all frames to a single cell, which prevents exploration).
To identify good downscaling parameters we define a downscaling objective function that estimates the quality of the balance achieved.
Because appropriate downscaling parameters can vary across Atari games and even as exploration progresses within a given game, such parameters are optimised \emph{dynamically} at regular intervals by maximising this objective function over a sample of recently discovered frames (Methods).

Here, the robustification phase consists of a modified version of the ``backward algorithm''\cite{salimans2018learning}, currently the highest performing LfD algorithm on \emph{Montezuma's Revenge}.
Due to the large computational expense of the robustification process, the focus of this work is on eleven games that have been considered hard-exploration by the community\cite{bellemare2016unifying} or for which the state-of-the-art performance was still below human performance (Methods ``\nameref{sec:atari_sota}''). 
To ensure that the trained policy becomes robust to environmental perturbations, during robustification stochasticity is added to these environments following current community standards for what constitutes the appropriately difficult version of this benchmark\cite{Machado2018RevisitingTA}.
The rewards in these games differ by orders of magnitude and standard reward clipping overemphasises small rewards over bigger payoffs.
However, the exploration phase provides extensive information about the rewards available in each game, making it possible to automatically scale rewards to an appropriate range (Methods).

\begin{figure}
    \centering
    \begin{subfigure}[b]{0.95\textwidth}
        \centering
        \includegraphics[width=\linewidth]{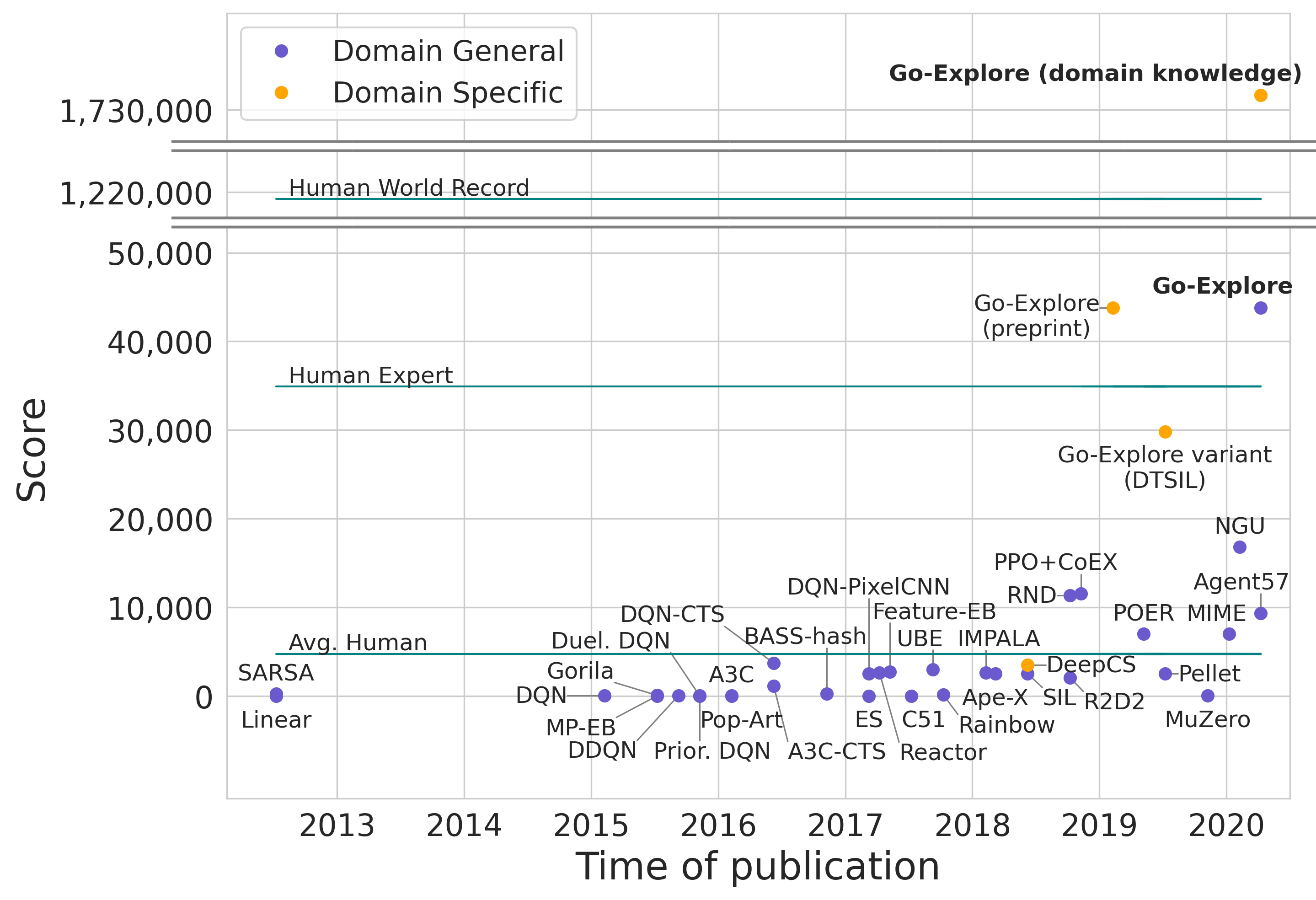}
        \caption{Historical progress on Montezuma's Revenge.}
        \label{fig:montezuma_hist}
        \vspace{5mm}
    \end{subfigure}
    \begin{subfigure}[b]{0.99\textwidth}
        \centering
        \fontsize{9}{10.5}\selectfont
        \begin{tabular}{ c | r | r r r } 
            Game & \shortstack{Exploration \\ Phase} & \shortstack{Robustification \\ Phase} & SOTA & Avg. Human \\
            \hline
            Berzerk & 131,216 & \textbf{197,376} & 1,383 & 2,630 \\
            Bowling & 247  & \textbf{260} & 69 & 160 \\
            Centipede & 613,815 & \textbf{1,422,628} & 10,166 & 12,017 \\
            Freeway & 34 & \textbf{34} & \textbf{34} & 30 \\
            Gravitar & 13,385 & \textbf{7,588} & 3,906 & 3,351 \\
            MontezumaRevenge & 24,758 & \textbf{43,791} & 11,618 & 4,753 \\ 
            Pitfall & 6,945 & \textbf{6,954} & 0 & 6,463 \\
            PrivateEye & 60,529 & \textbf{95,756} & 26,364 & 69,571 \\
            Skiing & -4,242 & \textbf{-3,660} & -10,386 & -4,336 \\
            Solaris & 20,306 & \textbf{19,671} & 3,282 & 12,326 \\
            Venture & 3,074 & \textbf{2,281} & 1,916 & 1,187 \\
            
        \end{tabular}
        \caption{Performance on the 10 focus games of the Go-Explore variant that uses the downscaled representation during the exploration phase. Bold indicates the best scores with stochastic evaluation. A video of high performing runs can be found at \href{https://youtu.be/e_aqRq59-Ns}{https://youtu.be/e\_aqRq59-Ns}.}
        \label{fig:phase2_sota}
    \end{subfigure}
    \caption{\textbf{Performance of robustified Go-Explore on Atari games.} (a) Go-Explore produces massive improvements over previous methods on Montezuma's Revenge, a grand challenge which was the focus of intense research for many years. (b) It exceeds the average human score in each of the 11 hard-exploration and unsolved games in the Atari suite, and matches or beats (often by a factor of 2 or more) the state of the art in each of these games. }
\end{figure}

At test time, the final mean performance of Go-Explore is both superhuman and surpasses the state of the art in all eleven games (except in Freeway where both Go-Explore and the state of the art reach the maximum score; Fig.~\ref{fig:phase2_sota}).
These games include the grand challenges of \emph{Montezuma's Revenge}, where Go-Explore quadruples the state-of-the-art score, and \emph{Pitfall}, where Go-Explore surpasses average human performance while previous algorithms were unable to score any points\footnote{
Recall that only results evaluated according to community-defined standards as to the appropriately challenging stochastic version of Atari are considered here, namely those with sticky actions\cite{Machado2018RevisitingTA}. Some algorithms have scored points on Pitfall without sticky actions, including recent concurrent work\cite{badia2020agent57,puigdomenech2020never}.}. Also noteworthy is the performance on \emph{Private Eye}, where Go-Explore is able to reliably achieve the highest possible score in 4 out of 5 runs, and the performance on \emph{Skiing}, where Go-Explore outperforms human performance despite the fact that the reward structure of this game makes it notoriously hard to train on\cite{such2017deep,badia2020agent57}.

Interestingly, there are many cases where the robustified policies obtain higher scores than the trajectories discovered during the exploration phase.
There are two reasons for this:
first, while the exploration phase does optimise for score by updating trajectories to higher scoring ones, the robustification phase
is more effective at fine-grained reward optimisation;
second, we provide the backward algorithm with demonstrations from multiple (10) runs of the exploration phase, thus allowing it to follow the best trajectory in the sample while still benefiting from the data contained in worse trajectories.

\begin{figure}
    \centering
    \includegraphics[width=\linewidth]{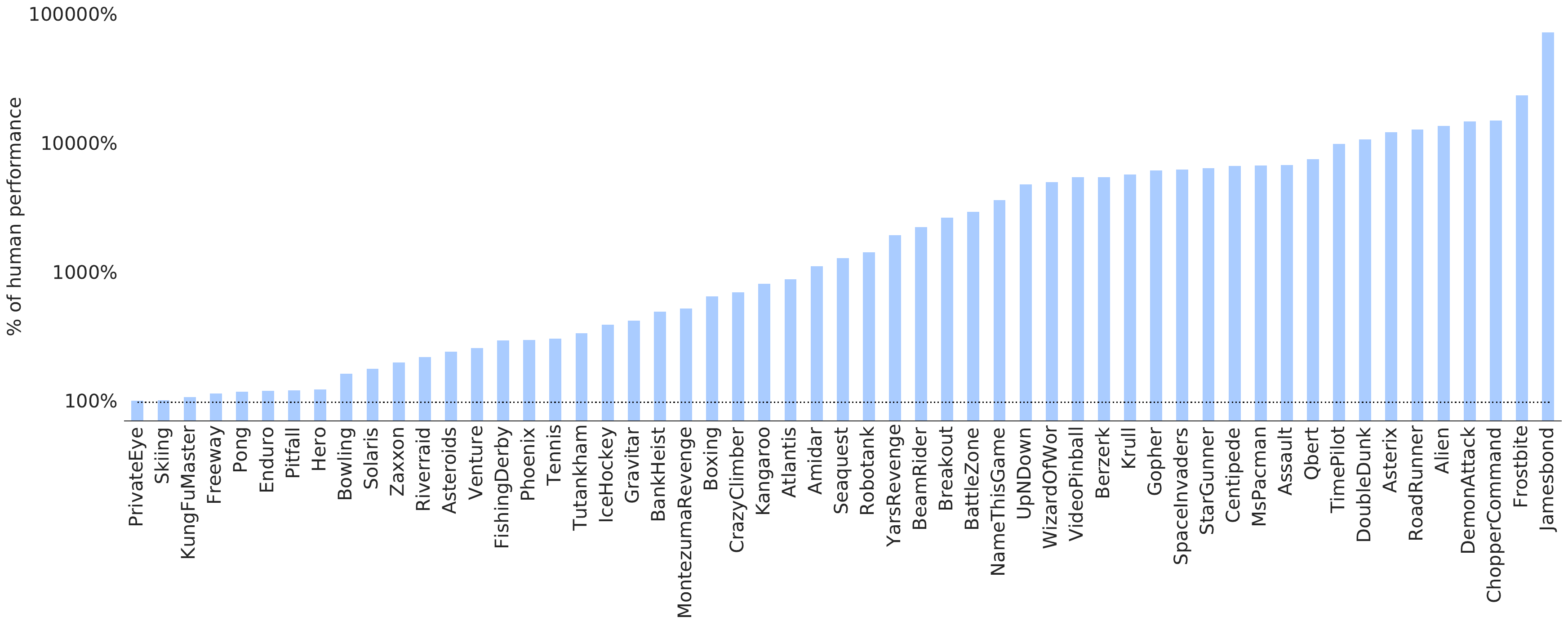}
    \caption{\textbf{Human-normalised performance of the exploration phase on all Atari games.} The exploration phase of Go-Explore exceeds human performance in every game, often by orders of magnitude. 
    }
    \label{fig:phase1_sota}
\end{figure}

The ability of the Go-Explore exploration phase to find high-performing trajectories is not limited to hard-exploration environments; it finds trajectories with superhuman scores for all of the 55 Atari games provided by OpenAI gym\cite{brockman}, a feat that has not been performed before (save concurrent work published just prior to this work \cite{badia2020agent57})
and in 83.6\% of these games the trajectories reach scores higher than those achieved by state-of-the-art algorithms (Fig.~\ref{fig:phase1_sota}), suggesting its potential for automatically generating solution demonstrations in many other challenging domains.
Modern RL algorithms are already able to adequately solve the games not included in the 11 focus games in this work, as demonstrated by previous work (Extended Data Table~\ref{etab:atari_full_scores}).
Thus, because robustifying these already solved games would have been prohibitively expensive, we did not perform robustification experiments for these 44 games.

Go-Explore can also harness easy-to-provide domain knowledge to substantially boost performance.
It can do so because the way Go-Explore explores a state space depends on the cell representation, meaning exploration efficiency can be improved by constructing a cell representation that only contains features relevant for exploration.
We demonstrate this principle in the games of Montezuma's Revenge and Pitfall, where the features are the agent's room number and coordinates for both games, as well as the level number and keys for Montezuma's Revenge.

The exploration phase with added domain knowledge explores both games extensively, discovering all 255 rooms of Pitfall and completely solving Montezuma's Revenge by reaching the end of level 3.
After robustification, the resulting policies achieve a mean score of 102,571 on Pitfall, close to the maximum achievable score of 112,000. 
On Montezuma's Revenge, the mean score of the robustified policies is 1,731,645, 
exceeding the previous state of the art score by a factor of 150.
This post-robustification score on Montezuma's Revenge is not only much greater than average human performance, but in fact is greater than the human world record of 1.2 million\cite{atari_scoreboard} (Fig.~\ref{fig:montezuma_hist}).
Notably, the performance on Montezuma's Revenge is limited not by the quality of the learned agents, but by the default time limit of 400,000 frames that is imposed by OpenAI Gym, but is not inherent to the game.
With this limit removed, robustified agents appear to be able to achieve
arbitrarily high scores, with one agent frequently reaching a score
of over 40 million after 12.5 million frames (the equivalent of about 58 hours of continuous game play).

Besides effectively representing all information relevant for exploration, the spatial cell representation enables improving cell selection during exploration, such as by considering whether neighbouring cells have been visited. Doing so greatly improves performance in Montezuma's Revenge, though it has little effect on Pitfall.

\section*{A hard-exploration robotics environment}
\label{sec:robotics}

While robotics is a promising application for reinforcement learning and
it is often easy to define the goal of a robotics task at a high level (e.g.\ to put a cup in the cupboard), it is much more difficult to define an appropriate and sufficiently dense reward function (which would need to consider all of the low-level motor commands involved in opening the cupboard, moving toward the cup, grasping the cup, etc.).
Go-Explore allows us to forgo such a dense reward function in favor of a sparse reward function that only considers the high-level task.
In addition, robot policies are usually trained in simulation before being transferred to the real world\cite{cully2015robots,andrychowicz2020learning,peng2018sim,tan2018sim}, making robotics a natural domain to demonstrate the usefulness of harnessing the ability of simulators to restore states in reinforcement learning.

\begin{figure}
    \centering
    \begin{tabular}{c}
        \begin{subfigure}[t]{\textwidth}
            \centering
            \begin{tabular}{cccc}
                \includegraphics[width=0.2\linewidth]{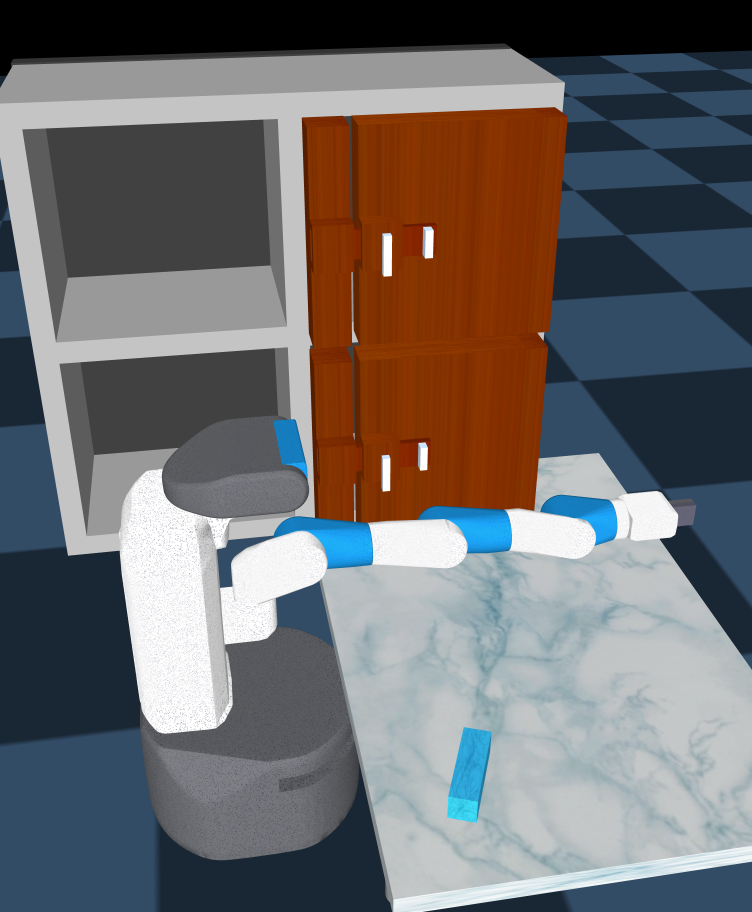} & \includegraphics[width=0.2\linewidth]{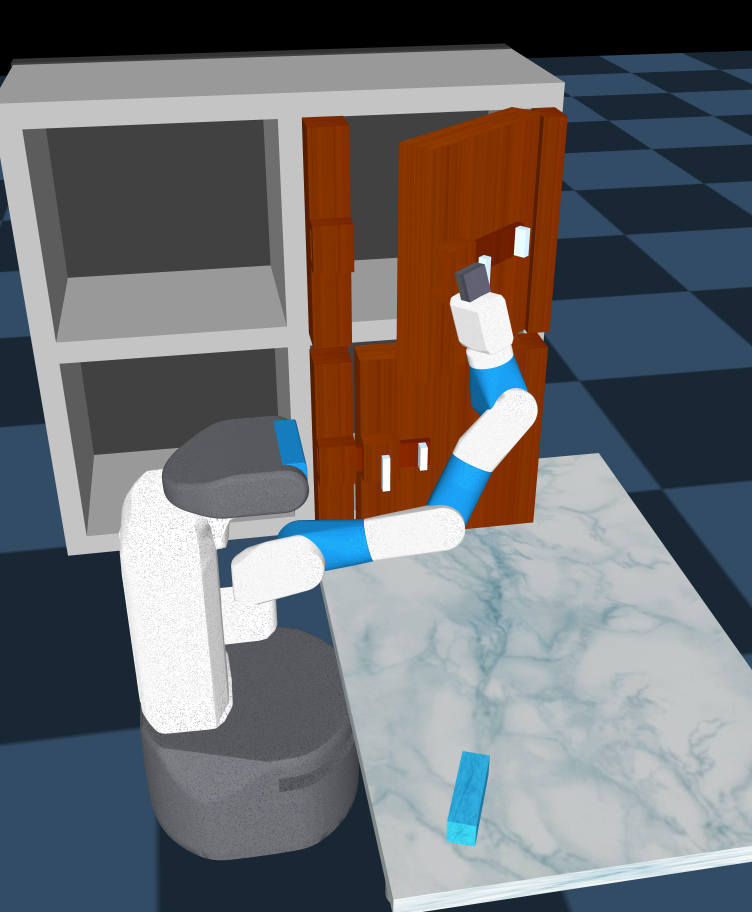} &
                \includegraphics[width=0.2\linewidth]{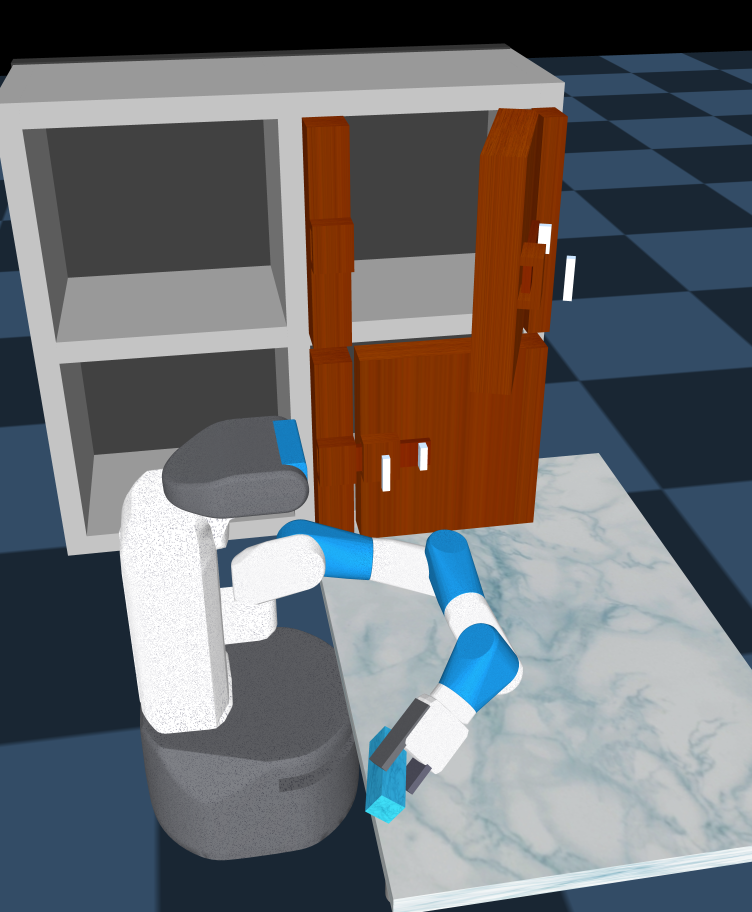} & \includegraphics[width=0.2\linewidth]{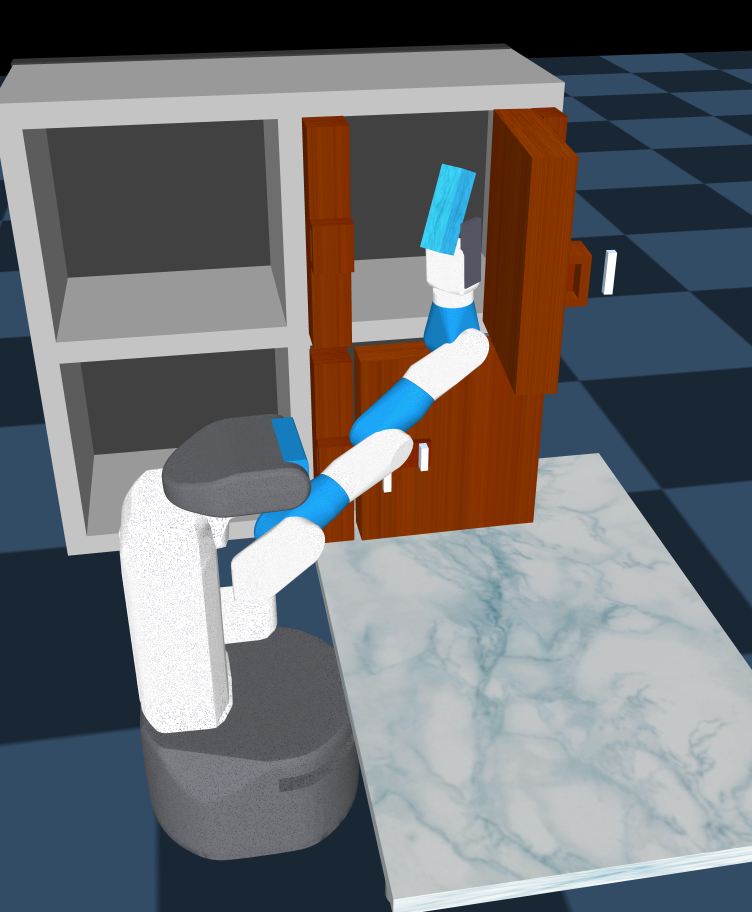}
            \end{tabular}
            \caption{Solving the top right shelf of the robotics environment.}
            \label{fig:robotics_picture}
            \vspace{2mm}
        \end{subfigure}
         \\
         \centering
         \begin{tabular}{cc}
            \begin{subfigure}[t]{0.45\linewidth}
                \centering
                \includegraphics[width=\linewidth]{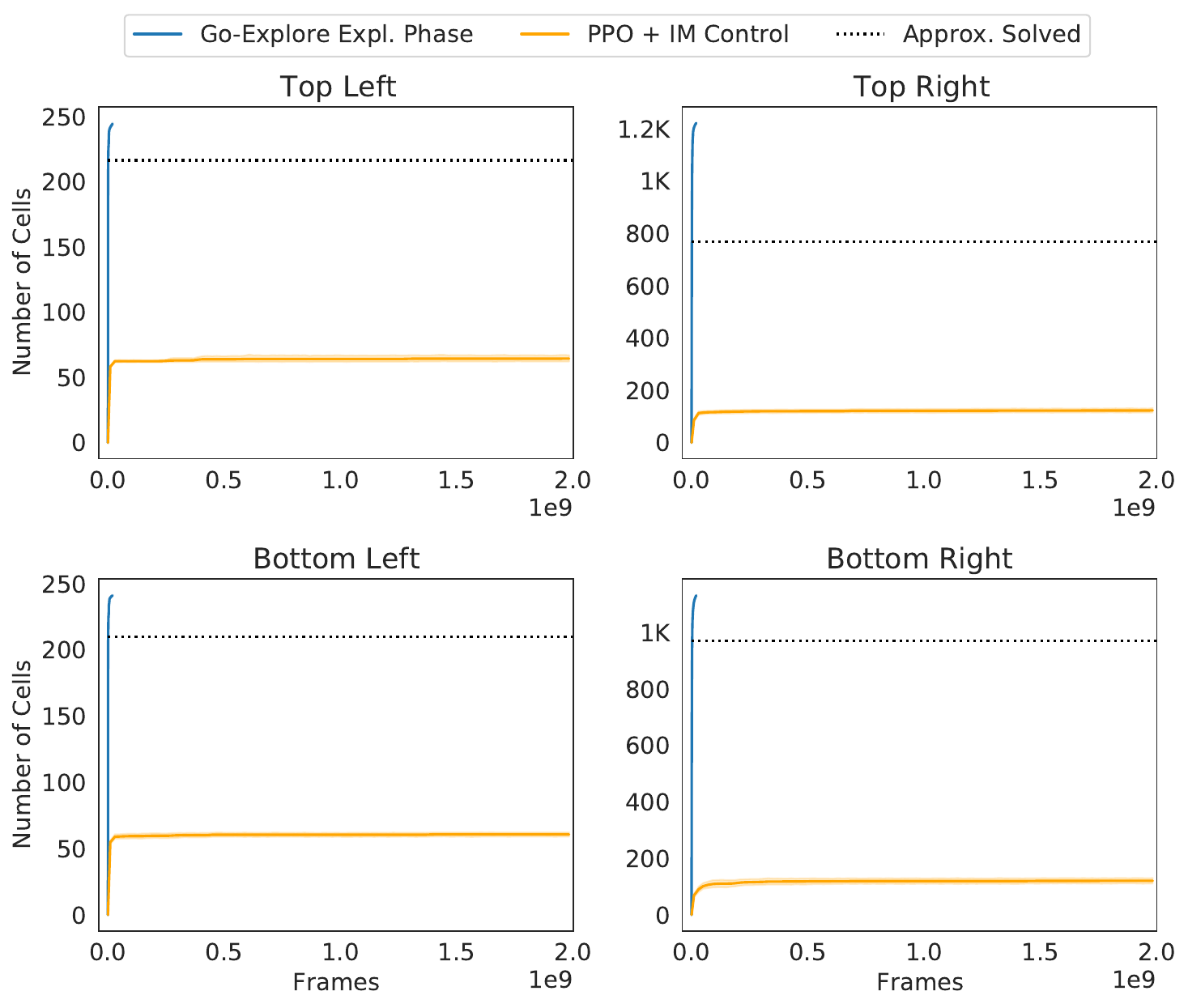}
                \caption{Cells discovered by the exploration phase and an intrinsic motivation control.}
                \label{fig:robotics_phase1_vs_control}
            \end{subfigure} 
            &
            \centering
            \begin{subfigure}[t]{0.45\linewidth}
                \centering
                \includegraphics[width=\linewidth]{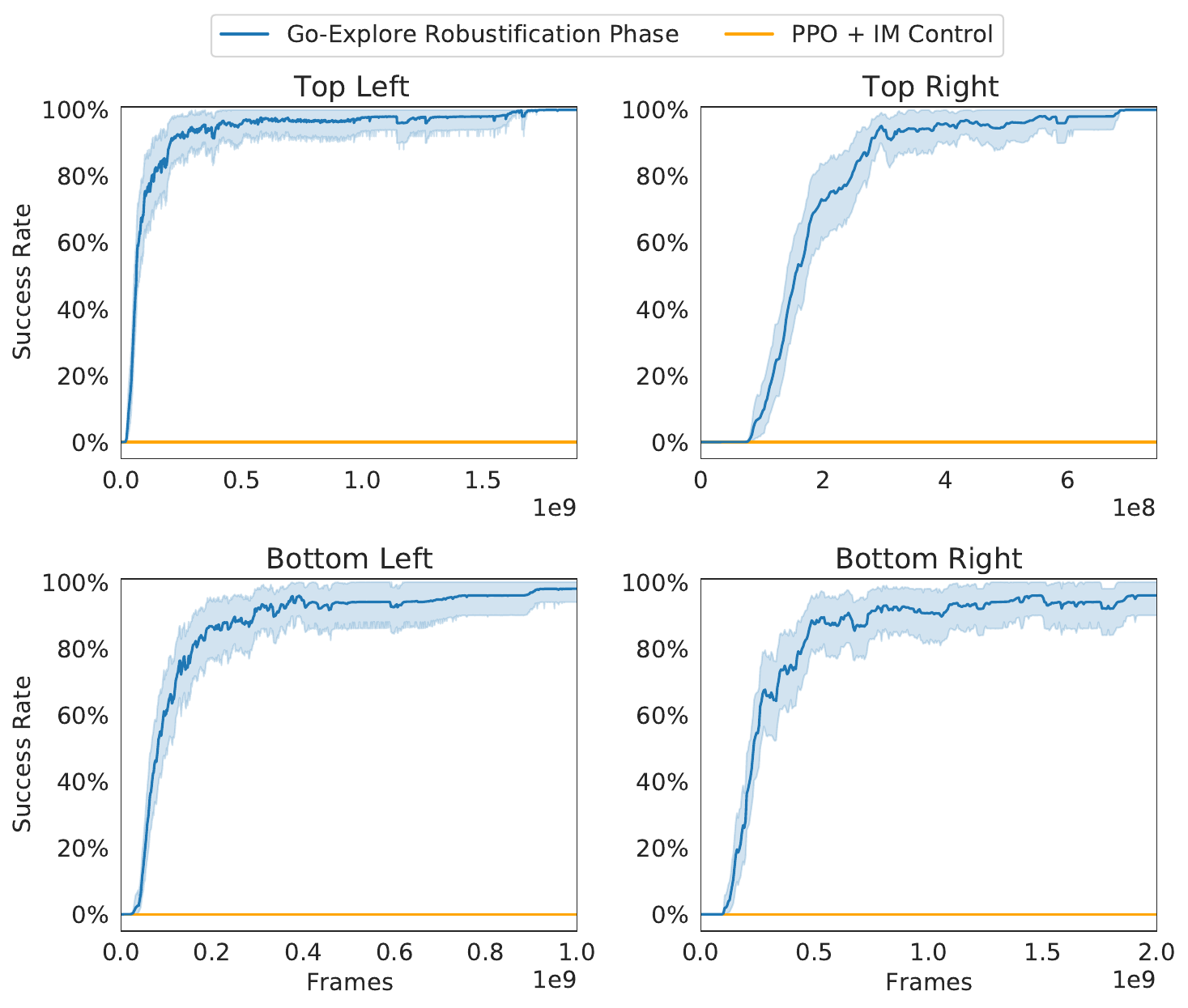}
                \caption{Robustification progress per target shelf.}
                \label{fig:robotics_robustification}
            \end{subfigure}
         \end{tabular}
    \end{tabular}
    \caption{\textbf{Go-Explore can Solve a Challenging, Sparse-Reward, Simulated Robotics Task.}  (a) A simulated Fetch robot needs to grasp an object and put it in one of four shelves. (b) The exploration phase significantly outperforms an intrinsic motivation control using the same cell representation. (c) For each of four different target locations, including the two with a door, the robot is able to learn to pick the object up and place it on the shelf in 99\% of trials.}
    \label{fig:robot}
\end{figure}

Accordingly, the following experiment, performed in a simulated environment, demonstrates that Go-Explore can solve a practical hard-exploration task in which a robot arm needs to pick up an object and put it inside of one of four shelves (Fig.~\ref{fig:robotics_picture}).
Two of the shelves are behind latched doors, providing an additional challenge.
A reward is given \emph{only} when the object is put into a specified target shelf.
Grasping an object and picking it up is itself a challenging problem in robotics, in part due to the challenge of exploration\cite{nair2018overcoming,kraft2010development}. Roboticists are often forced to employ the imperfect workaround of simplifying exploration by restricting the robot's range of motion substantially to favour arm positions that make grasping more likely\cite{andrychowicz2017hindsight}.
In our setup, by contrast, all nine joints of the robot arm are independently controlled to the full extent permitted by the robot's specifications.
A state-of-the-art RL algorithm for continuous control (PPO\cite{Schulman2017ProximalPO}) does not encounter a single reward after training in this environment for a billion frames, showcasing the hard-exploration nature of this problem.
For this problem, Go-Explore returns by restoring simulator state, explores by taking random actions, and assigns states to cells with an easy-to-provide domain-knowledge-based mapping (Methods).

In the exploration phase, Go-Explore quickly and reliably discovers successful trajectories for putting the object in each of the four shelves (Fig.~\ref{fig:robotics_phase1_vs_control}).
In contrast, a count-based intrinsic motivation algorithm with the same domain knowledge representation as the exploration phase is incapable of discovering any reward (Fig.~\ref{fig:robotics_robustification}), and indeed discovers only a fraction of the cells discovered by the exploration phase after 1 billion frames of training, 50 times more than the exploration phase of Go-Explore (Fig.~\ref{fig:robotics_phase1_vs_control}). In spite of receiving intrinsic rewards for touching the object, this control was incapable of learning to reliably grasp, thus making a large fraction of cells unreachable (including those necessary to obtain rewards). Evidence suggests that this failure to grasp is due to the problem of \emph{derailment}, which Go-Explore is specifically designed to solve (Supplementary Information).

While only a single trajectory was extracted for robustification from each exploration-phase
run of Go-Explore in the Atari experiments, the trajectories discovered in the robotics task are exceptionally diverse (e.g.\ some throw the object into the shelf while others gently deposit it), making it possible to provide additional information to the robustification phase by extracting multiple trajectories from each exploration-phase run.
Robustifying these trajectories produces robust and reliable policies in $99\%$ of cases (Fig.~\ref{fig:robotics_robustification}).

\section*{Policy-based Go-Explore}
\label{sec:policy}

Leveraging the ability of simulators to restore states provides many benefits that increase the efficiency of Go-Explore, but it is not a requirement.
When returning, instead of restoring simulator state, it is possible to execute a policy conditioned on (i.e.\ told to go to) the state or cell to return to, which we call \emph{policy-based Go-Explore}. 
While returning to a state in this manner is less efficient than restoring simulator state 
(the actions required to return to a state need to be executed in the environment), there are nevertheless advantages to doing so.
First, having access to a policy in the exploration phase makes it possible to sample from this policy during the explore step, which has the potential to substantially increase its efficiency vs.\ taking random actions thanks to the generalisation properties of the policy (e.g.\ the algorithm need only learn how to cross a specific obstacle once instead of solving that problem again via random actions each time).
Second, training a policy in the exploration phase obviates the need for robustification and thus removes its associated additional complexity, hyperparameters, and overhead (see `Implementation Details' in Methods).
Finally, in a stochastic environment, it may not always be possible to reliably repeat certain lucky sequences of transitions that enabled visiting a state, even when taking optimal actions.
While this issue did not prove to be a major roadblock in our experiments, in other domains it could prevent certain trajectories from being robustified. 
Exploring directly in the stochastic environment with the help of a policy 
makes it possible to detect these cases early and take appropriate measures to avoid them (Supplementary Information).

In policy-based Go-Explore, a goal-conditioned policy is trained during the exploration phase with a common reinforcement learning algorithm (PPO\cite{Schulman2017ProximalPO}).
Because goal-conditioned policies often struggle to reach far away states\cite{eysenbach2019search}, instead of training the policy to directly reach cells in the archive, the policy is trained to follow the best trajectory of cells that previously led to the selected state (Methods).
In addition, because trajectories that successfully overcome difficult obstacles are initially rare (the network has to rely on sequences of random actions to initially pass those obstacles) policy-based Go-Explore also performs self-imitation learning to extract as much information from these successful trajectories as possible\cite{Oh2018SelfImitationL}.
After our publication of an early pre-print presenting Go-Explore without a return policy, and while our work on policy-based Go-Explore was well underway, another team independently introduced a Go-Explore variant that is similarly policy-based\cite{guo2019efficient}; policy-based Go-Explore substantially outperforms this variant in terms of performance and sample efficiency (see Supplementary Information for a full comparison).

The goal-conditioned policy has the potential to greatly improve exploration effectiveness over random actions; in addition to returning to previously visited cells, the policy can also be queried during the explore step by presenting it with additional goals, including goals not already in the archive.
Such goal cells are chosen according to a simple strategy that either chooses a cell adjacent to the current position of the agent, possibly leading to a new cell, or randomly selects a cell from the archive, potentially repositioning the agent within the archive or finding new cells while trying to do so (Methods).
To examine whether this exploration strategy is indeed more effective than taking random actions, every time post-return exploration is started, the algorithm randomly commits with equal probability to either taking random actions or sampling from the policy for the duration of the post-return exploration step, thus ensuring that each strategy will be employed in similar scenarios equally often.

\begin{figure}
    \centering
    \begin{subfigure}[t]{0.45\textwidth}
        \centering
        \includegraphics[width=\linewidth]{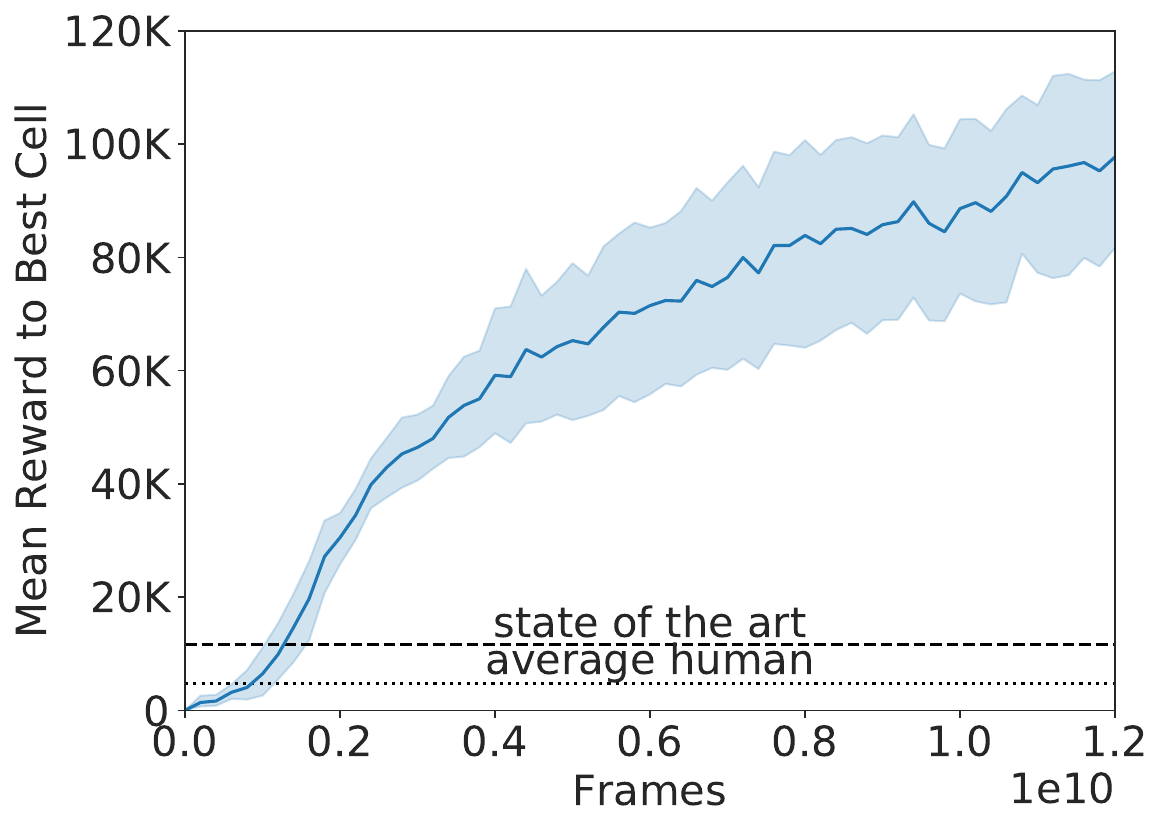}
        \caption{Montezuma's Revenge}
    \end{subfigure}%
    \begin{subfigure}[t]{0.45\textwidth}
        \centering
        \includegraphics[width=\linewidth]{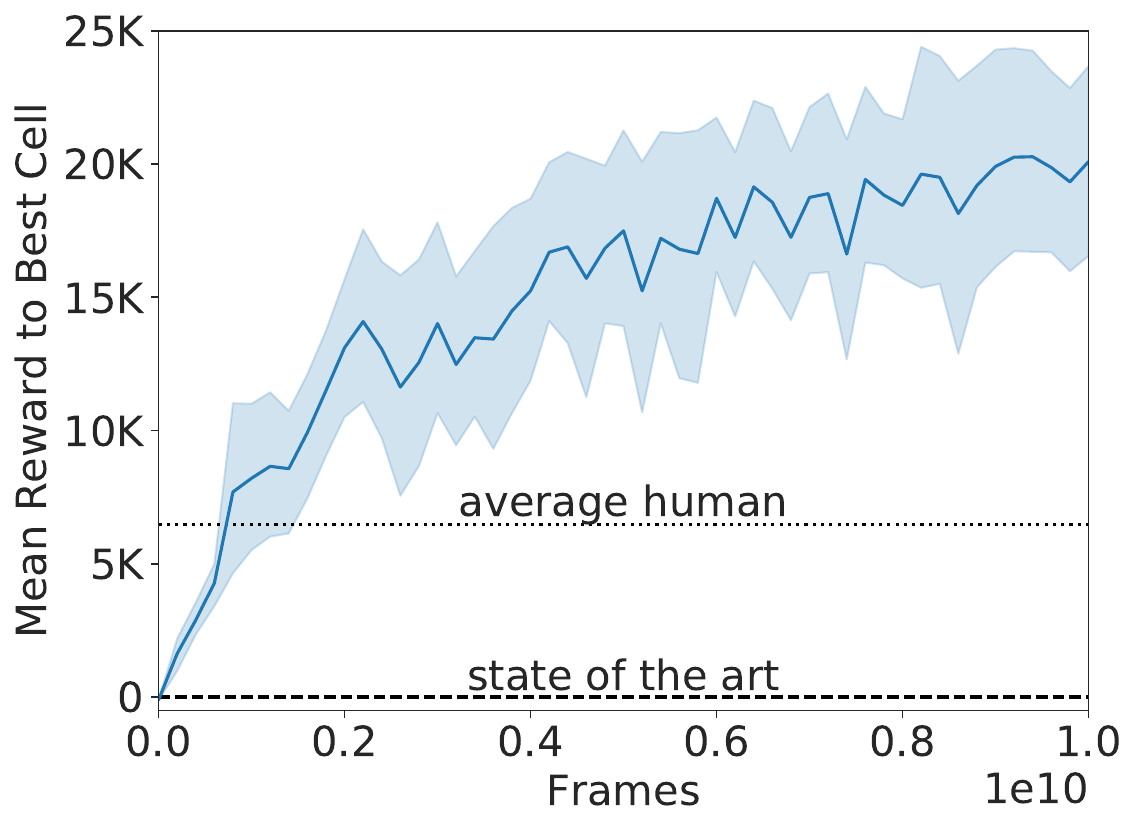}
        \caption{Pitfall}
    \end{subfigure}%
    \caption{\textbf{Policy-based Go-Explore with domain knowledge outperforms state-of-the-art and human performance in Montezuma's Revenge and Pitfall.} 
    On both Montezuma's Revenge \textbf{(a)} and Pitfall \textbf{(b)}, performance increases throughout the run, suggesting even higher performance is possible with additional training time.
    }
    \label{fig:policy_ge}
\end{figure}

Policy-based Go-Explore was tested on Montezuma's Revenge and Pitfall with the domain-knowledge cell representation.
On these games, it beats the state of the art and human performance with a mean reward of 97,728 points on Montezuma's Revenge and 20,093 points on Pitfall (Fig.~\ref{fig:policy_ge}), demonstrating that the performance of Go-Explore is not merely a result of its ability to leverage simulator restorability, but a function of its overall design.
Furthermore, we found that sampling from the policy is more effective at discovering new cells than taking random actions, and becomes increasingly more effective across training as the policy gains new, generally useful skills,
ultimately resulting in the discovery of over four times more cells than random actions on both Montezuma's Revenge and Pitfall (Extended Data Fig.~\ref{efig:policy_based_ge_cell_discovery}).
Given that random actions form the basis of exploration for many RL algorithms, this result highlights the potential for improvement by moving towards goal-conditioned policy-based exploration.

\section*{Conclusion}
\label{sec:conclusion}

The effectiveness of Go-Explore on the problems presented in this work suggest that it will enable progress in many domains that can be framed as sequential decision making problems, including robotics\cite{cully2015robots,andrychowicz2020learning,peng2018sim,tan2018sim}, neural architecture search\cite{liu2018darts}, language understanding\cite{madotto2020exploration}, scheduling\cite{zhang1995reinforcement}, marketing\cite{abe2004cross}, and drug design\cite{popova2018deep}.
In addition, as a result of the modular nature of the Go-Explore family, the instantiations described in this work are 
only a fraction of the possible ways in which the Go-Explore paradigm can be implemented.
Other instantiations may learn the cell representation, learn to choose which cells to return to, learn which cells to try to reach during the exploration step, learn a separate, specialised policy whose job is only to explore after returning, learn to explore safely in the real world by mining diverse catastrophes in simulation, maintain a continuous density-based archive rather than a discrete cell-based one,
leverage all transitions gathered during exploration for robustification,
and so on.
These unexplored variants could further improve the generality, performance, robustness, and efficiency of Go-Explore.
Finally, the insights presented in this work extend beyond Go-Explore; the simple decomposition of remembering previously found states, returning to them and then exploring from them appears to be especially powerful, suggesting it may be fundamental feature of learning in general.
Harnessing these insights, either within or outside of the context of Go-Explore, may be a key to improving our ability to create generally intelligent agents.

\printbibliography[segment=1,check=onlynew]
\endrefsegment
\newrefsegment

\clearpage

\begin{methods}

\subsection{State of the art on Atari}
\label{sec:atari_sota}

With new work on RL for Atari being published on a regular basis, and with reporting methods often varying significantly, it can be difficult to establish the state-of-the-art score for each Atari game. At the same time, it is important to compare new algorithms with previous ones to evaluate progress.

For determining the state-of-the-art score for each game, we considered a set of notable, recently published papers that cover at least the particular subset of games this paper focuses on, namely hard-exploration games. Community guidelines advocate \emph{sticky actions} as a way to evaluate agents on Atari\cite{Machado2018RevisitingTA}, and there is substantial evidence to show that sticky actions can decrease performance substantially compared to the now deprecated \emph{no-ops} evaluation strategy\cite{Machado2018RevisitingTA,castro18dopamine,toromanoff2019deep}. As a result, we exclude work which was only evaluated with no-ops from our definition of state of the art. Fig.~\ref{fig:montezuma_hist} includes works tested only with no-ops as they help bring context to the amount of effort expended by the community on solving Montezuma's Revenge. We did not include work that does not provide individualised scores for each game. To avoid cherry-picking lucky rollouts that can bias scores upward substantially, we also exclude work that only provided the maximum score achieved in an entire run as opposed to the average score achieved by a particular instance of the agent. 

In total, state of the art results were extracted from the following papers: \citefull{burda:rnd2018}, \citefull{castro18dopamine}, \citefull{Choi2018ContingencyAwareEI}, \citefull{fedus2019hyperbolic}, \citefull{Taiga2020On}, \citefull{tang2020taylor}, and \citefull{toromanoff2019deep}.
Because these works themselves report scores for several algorithms and variants, including reproductions of previous algorithms, a total of 23 algorithms and variants were included in the state of the art assessment. 
For each game, the state-of-the-art score was the highest score achieved across all algorithms.

\subsection{Downscaling on Atari}

In the first variant of Go-Explore presented in this work (Sec.~ ``\nameref{sec:deterministic}''), the cell representation is a downscaled version of the original game frame. To obtain the downscaled representation, (1) the original frame is converted to grayscale, (2) its resolution is reduced with pixel area relation interpolation to a width $w \le 160$ and a height $h \le 210$, and (3) the pixel depth is reduced to $d \le 255$ using the formula $\lfloor \frac{d \cdot p}{255} \rfloor$, where $p$ is the value of the pixel after step (2).
The parameters $w$, $h$ and $d$ are updated dynamically by proposing different values for each, calculating how a sample of recent frames would be grouped into cells under these proposed parameters, and then selecting the values that result in the best cell distribution (as determined by the objective function defined below).

The objective function for candidate downscaling parameters is calculated based on a target number of cells $T$ (where $T$ is a fixed fraction of the number of cells in the sample), the actual number of cells produced by the parameters currently considered $n$, and the distribution of sample frames over cells $\mathbf{p}$. Its general form is
\begin{equation}
O(\mathbf{p}, n) = \frac{H_n(\mathbf{p})}{L(n, T)} 
\end{equation}
$L(n, T)$ measures the discrepancy between the number of cells under the current parameters $n$ and the target number of cells $T$.
It prevents the representation that is discovered from aggregating too many frames together, which would result in low exploration, or from aggregating too few frames together, which would result in an intractable time and memory complexity, and is defined as
\begin{equation}
L(n, T) = \sqrt{\left| \frac{n}{T} - 1 \right| + 1}
\end{equation}
$H_n(\mathbf{p})$ is the ratio of the entropy of how frames were distributed across cells to the entropy of the discrete uniform distribution of size $n$, i.e.~the normalised entropy. In this way, the loss encourages frames to be distributed as uniformly as possible across cells, which is important because highly non-uniform distributions may suffer from the same lack of exploration that excessive aggregation can produce or the same intractability that lack of aggregation can produce. Unlike unnormalized entropy, normalised entropy is comparable across different numbers of cells, allowing the number of cells to be controlled solely by $L(n, T)$. Its form is
\begin{equation}
H_n(\mathbf{p}) = -\sum^n_{i=1} \frac{p_i \log(p_i)}{\log(n)}
\end{equation}
At each step of the randomised search, new values of each parameter $w$, $h$ and $d$ are proposed by sampling from a geometric distribution whose mean is the current best known value of the given parameter. If the current best known value is lower than a heuristic minimum mean (8 for $w$, 10.5 for $h$ and 12 for $d$), the heuristic minimum mean is used as the mean of the geometric distribution. New parameter values are re-sampled if they fall outside of the valid range for that parameter.

The recent frames that constitute the sample over which parameter search is done are obtained by maintaining a set of recently seen sample frames as Go-Explore runs: each time a frame not already in the set is seen during the explore step, it is added to the running set with a probability of 1\%. If the resulting set contains more than 10,000 frames, the oldest frame it contains is removed.

\subsection{Domain knowledge representations}

The domain knowledge representation for Pitfall consists of the current room (out of 255) the agent is currently located in, as well as the discretized $x,y$ position of the agent (Extended Data Table~\ref{etab:backward_hyper_atari}). In Montezuma's Revenge, the representation also includes the keys currently held by the agent (including which room they were found in) as well as the current level. Although this information can in principle be extracted from RAM, in this work it was extracted from pixels through small hand-written classifiers, showing that domain knowledge representations need not require access to the inner state of a simulator.

In robotics, the domain knowledge representation is extracted from the internal state of the MuJoCo\cite{todorov2012mujoco} simulator. However, similar information has been extracted from raw camera footage for real robots by previous work\cite{andrychowicz2020learning}. It consists of the current 3D position of the robot's gripper, discretized in voxels with sides of length 0.5 meters, whether the robot is currently touching (with a single grip) or grasping (touching with both grips) the object, and whether the object is currently in the target shelf. In the case of the two target shelves with doors, the positions of the door and its latch are also included. The discretization for latches and doors follows the following formula, given that $d$ is the distance of the latch/door from its starting position in meters: $\lfloor \frac{d + 0.195}{0.2}\rfloor$.

\subsection{Exploration phase}
\label{sec:exploration_phase}

During the exploration phase, the selection probability of a cell at each step is proportional to its \emph{selection weight}. For all treatments except Montezuma's Revenge with domain knowledge but without a return policy, the cell selection weight $W$ is listed in Extended Data Table~\ref{etab:hyper}.

In the case of Montezuma's Revenge with domain knowledge but without a return policy, three additional domain knowledge features further contribute to the weight: 
(1) The number of horizontal neighbours to the cell present in the archive ($h$); 
(2) A key bonus: for each location (defined by level, room, and x,y position), the cell with the largest number of keys at that location gets a bonus of $k = 1$ ($k = 0$ for other cells);
(3) the current level. The first two values contribute to the location weight
\begin{equation}
    W_\mathrm{location} = \frac{2 - h}{10} + k.
\end{equation}
This value is then combined with $W$ above as well as the level of the given cell $l$ and the maximum level in the archive $L$ to obtain the final weight for Montezuma's Revenge with domain knowledge:
\begin{equation}
    W_\mathrm{mont\_domain} = 0.1^{L - l}\left(W + W_\mathrm{location}\right).
\end{equation}
While it is possible to produce an analogous domain knowledge cell selection weight for Pitfall with domain knowledge, no such weight produced any substantial improvement over $W$ alone.

Unless otherwise specified, once a cell is returned to, exploration proceeds with random actions for a number of steps (100 in Atari, 30 in robotics), or until the end of episode signal is received from the environment. In Atari, where the action set is discrete, actions are chosen uniformly at random. In robotics, each of the 9 continuous-valued components of the action is sampled independently and uniformly from the interval from -1 to 1. To help explore in a consistent direction, the probability of repeating the previous action is 95\% for Atari and 90\% for robotics.

For increased efficiency, the exploration phase is processed in parallel by selecting a batch of return cells and exploring from each one of them across multiple processes. In all runs without a return policy, the batch size is 100.

\subsection{Evaluation}
\label{sec:evaluation}

In Atari, the score of an exploration phase run is measured as the highest score ever achieved at end of episode. This score is tracked by maintaining a virtual cell corresponding to the end of the episode.
An alternative approach would be to track the maximum score across all cells in the archive, regardless of whether they correspond to the end of episode, but this approach fails in games where rewards can be negative (e.g.~Skiing): in these cases, it is possible that the maximum scoring cell in the archive inevitably leads to future negative rewards, and is therefore not a good representation of the maximum score achievable in the game.
In practice, for games in which rewards are non-negative, the maximum score at end of episode is usually equal or close to the maximum score achieved in the entire archive.
For the 11 focus games, exploration phase scores are averaged across 50 exploration phase runs. For the other games, they are averaged across 5 runs. For domain knowledge, they are averaged across 100 runs.

\emergencystretch 3em  %
During robustification in Atari, a checkpoint is produced every 100 training iterations (13,926,400 frames).
A subset of checkpoints corresponding to points during which the rolling average of scores seen during training was at its highest are tested by averaging their scores across 100 test episodes.
Then the highest scoring checkpoint found is retested with 1,000 new test episodes to eliminate selection bias.
For the downscaled representation, robustification scores are averaged over 5 runs.
For domain knowledge, they are averaged across 10 runs. All testing is performed with sticky actions (see ``\nameref{sec:atari_sota}''). 
To accurately compare against the human world record of 1.2 million\cite{atari_scoreboard}, we patched an ALE bug that prevents the score from exceeding 1 million (Supplementary Information ``\nameref{sec:ale_issues}'').

The exploration phase for robotics was evaluated across 50 runs per target shelf, for a total of 200 runs. The reported metric is the proportion of runs that discovered a successful trajectory. Because the outcome of a robotics episode is binary (success or failure), there is no reason to continue robustification once the agent is reliably successful (unlike with Atari where it is usually possible to further improve the score).
Thus, robustification runs for robotics are terminated once they keep a success rate greater than 98.5\% for over 150 training iterations (19,660,800 frames), and the runs are then considered successful.
To ensure that the agent learns to keep the object inside of the shelf, a penalty of -1 is given for taking the object outside of the shelf, and during robustification the agent is given up to 54 additional steps after successfully putting the object in the shelf (see ``Extra frame coef'' in Extended Data Table~\ref{etab:backward_hyper_atari}), forcing it to ensure the object doesn't leave the shelf.
Out of 200 runs (50 per target shelf), 2 runs did not succeed after running for over 3 billion frames (whereas all other runs succeeded in fewer than 2 billion) and were thus considered unsuccessful (one for the bottom left shelf and the other for the bottom right shelf), resulting in a 99\% overall success rate. 

The robotics results are compared to two controls. First, to confirm the hard-exploration nature of the environment, 5 runs per target shelf of ordinary PPO\cite{Schulman2017ProximalPO} with no exploration mechanism were run for 1 billion frames. At no point during these runs were any rewards found, confirming that the robotics problem in this paper constitutes a hard-exploration challenge. Secondly, we ran 10 runs per target shelf for 2 billion frames of ordinary PPO augmented with count-based intrinsic rewards, one of the best modern versions of intrinsic motivation\cite{strehl2008analysis,tang2017exploration,bellemare2016unifying,Taiga2020On} designed to deal with hard-exploration challenges. The representation for this control is identical to the one used in the exploration phase, so as to provide a fair comparison. Similar to the exploration phase, the counts for each cell are incremented each time the agent enters a cell for the first time in an episode, and the intrinsic reward is given by $\frac{1}{\sqrt{n}}$, similar to $W$. Because it is possible (though rare) for the agent to place the object out of reach, a per-episode time limit is necessary to ensure that not too many training frames are wasted on such unrecoverable states. In robustification, the time limit is implicitly given by the length of the demonstration combined with the additional time described above and in Extended Data Table~\ref{etab:backward_hyper_atari}. For the controls, a limit of 300 time steps was given as it provides ample time to solve the environment (Extended Data Figure~\ref{efig:phase1_robotics_length}), while ensuring that the object is almost always in range of the robot arm throughout training. As shown in Fig.~\ref{fig:robotics_phase1_vs_control}, this control was unable to find anywhere near the number of cells found by the exploration phase, despite of running for significantly longer, and as shown in Fig.~\ref{fig:robotics_robustification}, it also was unable to find any rewards in spite of running for longer than any successful Go-Explore run (counting both the exploration phase and robustification phase combined).

\subsection{Hyperparameters}
\label{sec:hyperparameters}

Hyperparameters are reported in Extended Data Table~\ref{etab:hyper}. Extended Data Table~\ref{etab:atari_hyper} reports the hyperparameters specific to the Atari environment. Of note are the use of sticky actions as recommended by \citefull{Machado2018RevisitingTA}, and the fact that the agent acts every 4 frames, as is typical in RL for Atari\cite{mnih:nature15}. In this work, sample complexity is always reported in terms of raw Atari frames, so that the number of actions can be obtained by dividing by 4. In robotics, the agent acts 12.5 times per second. Each action is simulated with a timestep granularity of 0.001 seconds, corresponding to 80 simulator steps for every action taken. 

While the robustification algorithm originates from \citefull{salimans2018learning}, it was modified in various ways
(Supplementary Information ``\nameref{sec:backward}''). Extended Data Table~\ref{etab:backward_hyper_atari} shows the hyperparameters for this algorithm used in this work, to the extent that they are different from those in the original paper, or were added due to the modifications in this work. Extended Data Tables~\ref{etab:robotics_pos} and~\ref{etab:robotics_interest} show the state representation for robotics robustification. 

With the downscaled representation on Atari, the exploration phase was run for 2 billion frames prior to extracting demonstrations for robustification. Because exploration phase performance was slightly below human on Pitfall, Skiing and Private Eye, the exploration phase was allowed to run longer on these three games (5 billion for Pitfall and Skiing, 15 billion for Private Eye) to demonstrate that it can exceed human performance on all Atari games. The demonstrations used to robustify these three games were still extracted after 2 billion frames, and the robustified policies still exceeded average human performance thanks to the ability of robustification to improve upon demonstration performance. With the domain knowledge representation on Atari, the exploration phase ran for 1 billion frames. Robustification ran for 10 billion frames on all Atari games except Solaris (20 billion) and Pitfall when using domain knowledge demonstrations (15 billion). On robotics, the exploration phase ran for 20 million frames and details for the robustification phase are given in the Evaluation section.

\subsection{Policy-based Go-Explore}
\label{sec:policy_based}

The idea in policy-based Go-Explore is to learn how to return (rather than to restore archived simulator states to return). The algorithm builds off the popular PPO algorithm\cite{Schulman2017ProximalPO} described in Supplementary Information.
At the heart of policy-based Go-Explore lies a goal-conditioned policy $\pi_\theta(a| s, g)$ (Extended Data Fig.~\ref{efig:policy_based_ge_overview}), parameterized by $\theta$, that takes a state $s$ and a goal $g$ and defines a probability distribution over actions $a$.
Policy-based Go-Explore includes all PPO loss functions described in Supplementary Information, except that instances of the state $s$ are replaced with the state-goal tuple $(s, g)$.
The total reward $r_t$ is the sum of the trajectory reward $r^{\tau}_t$ (defined below) and the environment reward $r^e_t$, where $r^e_t$ is clipped to the $[-2, 2]$ range. 
Policy-based Go-Explore also includes self-imitation learning (SIL)\cite{Oh2018SelfImitationL} (Supplementary Information), where SIL actors follow the same procedure as regular actors, except that they replay the trajectory associated with the cell they select from the archive.
Hyperparameters are listed in Extended Data Table~\ref{etab:backward_hyper_atari}.

To fit the batch-oriented paradigm, policy-based Go-Explore updates its archive after every mini-batch (Extended Data Fig.~\ref{efig:policy_based_ge_overview}).
In addition, the ``go'' step now involves executing actions in the environment (as explained below), and each actor 
independently tracks whether it is in the ``go'' step or the ``explore'' step of the algorithm. 
For the purpose of updating the archive, no distinction is made between data gathered during the ``go'' step and data gathered during the ``explore'' step, meaning policy-based Go-Explore can discover new cells or update existing cells while returning.

For the experiments presented in this paper, data is gathered in episodes.
Whenever an actor starts a new episode, it selects a state from the archive according to the Go-Explore selection procedure and starts executing the ``go'' step.
Here, policy-based Go-Explore relies on its goal-conditioned policy to reach the selected state, which enables it to be applied without assuming access to a deterministic or restorable environment during the exploration phase.
It is exceedingly difficult and practically unnecessary to reach a particular state exactly, so instead, the policy is conditioned to reach the \emph{cell} associated with this state, referred to as the \emph{goal cell}, provided to the policy in the form of a concatenated one-hot encoding for every attribute characterizing the cell.
Directly providing the goal cell to the goal-conditioned policy did not perform well in preliminary experiments, presumably because goal-conditioned policies tend to falter when goals become distant\cite{eysenbach2019search}.
Instead, the actor is iteratively conditioned on the successive cells traversed by the archived trajectory that leads to the goal cell.

Here, we allow the agent to follow the archived trajectory in a soft-order, a method similar to the one described in \citefull{guo2019efficient}. 
To prevent the soft-trajectory from being affected by the time an agent spends in a cell, the algorithm first constructs a trajectory of non-repeated cells, collapsing any consecutive sequence of identical cells into a single cell.
Then, given a window size $N_w=10$, if the agent is supposed to reach a specific goal cell in this trajectory and it reaches that or any of the subsequent 9 cells in this trajectory, the goal is considered met.
When a goal is met, the agent receives a trajectory reward $r^{\tau}_t$ of 1 and the subsequent goal in the non-repeated trajectory (i.e. the goal that comes after the cell that was actually reached) is set as the next goal.
When the cell that was reached occurs multiple times in the window (indicating cycles) the next goal is the one that follows the last occurrence of this repeated goal cell.

As soon as an agent reaches the last cell in the trajectory, it receives a trajectory reward $r^{\tau}_t$ of 3 and executes the ``explore'' step, either through \emph{policy exploration} or \emph{random exploration}.
With policy exploration, the agent will select a goal for the policy according to one of three rules: (1) with $10\%$ probability, randomly select an adjacent cell (see ``Exploration phase'') not in the archive, (2) with $22.5\%$ probability, select any adjacent cell, whether already in the archive or not, and (3), in the remaining $67.5\%$ of cases, select a cell from the archive according to the standard cell-selection weights.
If the first rule does not apply because all adjacent cells are already in the archive, rules 2 and 3 are selected with proportionally scaled probabilities.
Note that, in the exploration step, the agent is presented directly with the goal, rather than with a trajectory.
Whenever the current exploration goal is reached, or if the goal is not reached for some number of steps (here 100), a new exploration goal is chosen. 
With random exploration, the agent takes random actions according to the random exploration procedure described in Methods ``\nameref{sec:exploration_phase}''.
All gathered data is ignored with respect to calculating the loss of the policy.

While following a trajectory or during exploration, it is possible for the agent to fail to make progress towards the current goal cell because the policy has converged towards putting all its probability mass on a small set of actions, meaning the policy performs insufficient exploration to discover the goal and observe its reward.
To alleviate this issue, in addition to having the entropy bonus $\mathcal{L}^{ENT}$, the policy is extended with an entropy term $e_t$ that divides the logits of the policy right before the softmax activation function is applied.
If the agent fails to reach the current goal for some number of steps $e^T_t$ (defined below), this entropy term is increased following:
\begin{equation}
    e_t(\hat{t}) = 1 + (\mathrm{max}(0, \hat{t} - e^T_t) \cdot e_f) ^ {e_p}
\end{equation}
where $\hat{t}$ is the number of steps the agent has taken since it last reached a goal (for returning) or discovered a new cell (for exploring), $e_f=0.01$ is the entropy increase factor and $e_p=2$ is the entropy increase power.
While executing the ``explore'' step, the threshold $e^T_t$ has a fixed value of 50. 
While returning, the threshold $e^T_t$ equals the number of actions that the followed trajectory required to move from the previously reached goal cell to the current goal cell.
Here, the previously reached goal cell refers to the first cell in the soft-trajectory window that matched the cell occupied by the agent at the time the previous goal was considered met.

Lastly, to prevent actors from spending many time steps without making any progress (possibly because the agent reached a state from which further progress is impossible), we terminate the episode early if the current goal is not reached within 1,000 steps after we have started to increase entropy (while returning), or if no new cells are discovered for 1,000 steps (while exploring). 
For Montezuma's Revenge with policy-based Go-Explore only, we also terminate the episode on death to deal with an ALE bug (details in Supplemental Information).

\subsection{Data availability} The data that support the findings of this study are available from the corresponding authors upon reasonable request.

\subsection{Code availability} The Go-Explore code is available at: \href{https://github.com/uber-research/go-explore}{https://github.com/uber-research/go-explore}.

\printbibliography[segment=2,check=onlynew]

\end{methods}

\begin{addendum}
\item[Acknowledgements]
We thank Ashley Edwards, Sanyam Kapoor, Felipe Petroski Such and Jiale Zhi for their ideas, feedback, technical support, and work on aspects of Go-Explore not presented in this work. We are grateful to the Colorado Data Center and OpusStack Teams at Uber for providing our computing platform. We thank Vikash Kumar for creating the MuJoCo files that served as the basis for or robotics environment  (\href{https://github.com/vikashplus/fetch}{https://github.com/vikashplus/fetch}).
\ifarxiv
\else

\fi

\item[Correspondence] should be addressed to Adrien Ecoffet (email: adrien.ecoffet@gmail.com), Joost Huizinga (email: joost.hui@gmail.com), and Jeff Clune (email: jclune@gmail.com).
\end{addendum}

\newpage

\section*{Extended Data}
\FloatBarrier

\renewcommand{\figurename}{Extended Data Figure}
\renewcommand{\tablename}{Extended Data Table}
\setcounter{figure}{0}
\setcounter{table}{0}

\begin{figure}[ht!]
    \begin{subfigure}[]{\textwidth}
            \centering
            \includegraphics[width=0.95\linewidth]{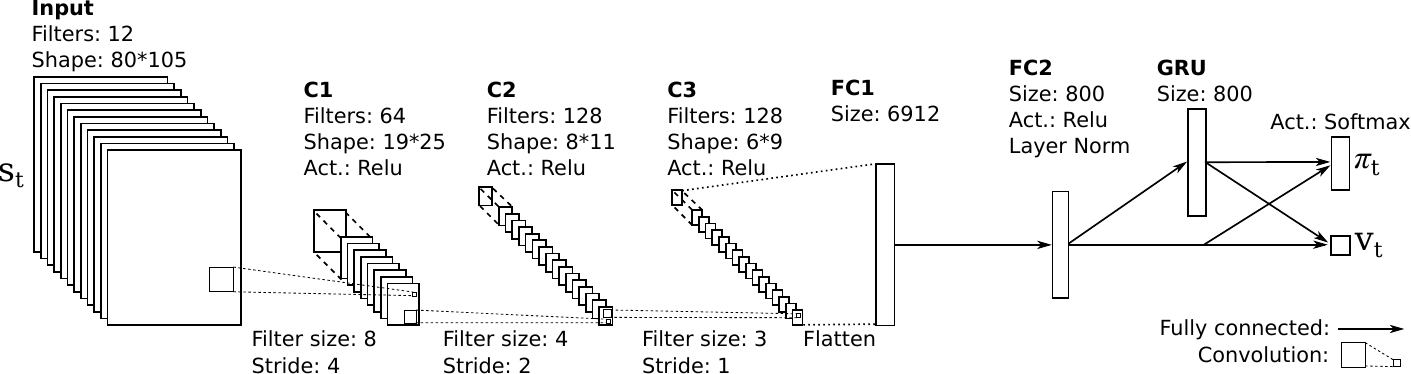}
            \caption{Atari architecture.}
        \label{efig:nn_atari}
    \end{subfigure}
    \begin{subfigure}[]{\textwidth}
            \centering
            \includegraphics[width=0.95\linewidth]{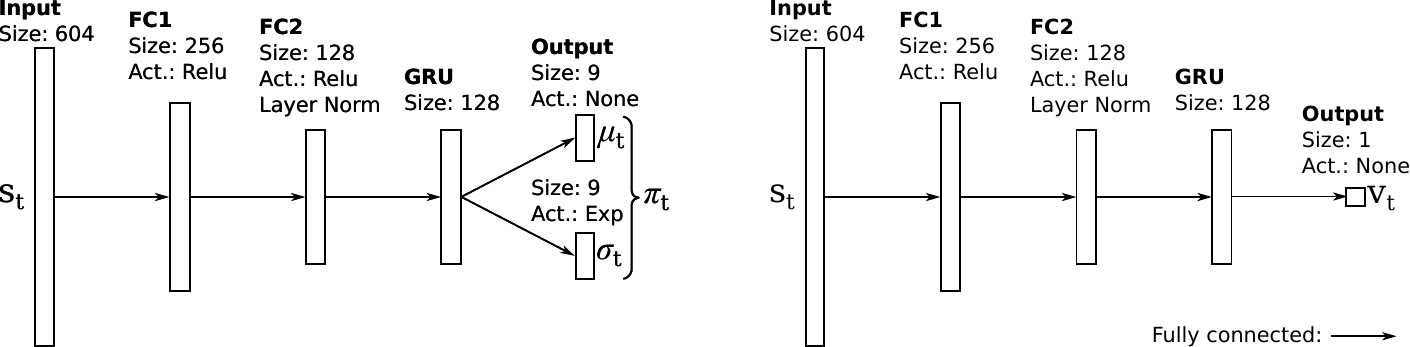}
            \caption{Robotics architecture.}
        \label{efig:nn_robotics}
    \end{subfigure}
    \begin{subfigure}[]{\textwidth}
            \centering
            \includegraphics[width=0.95\linewidth]{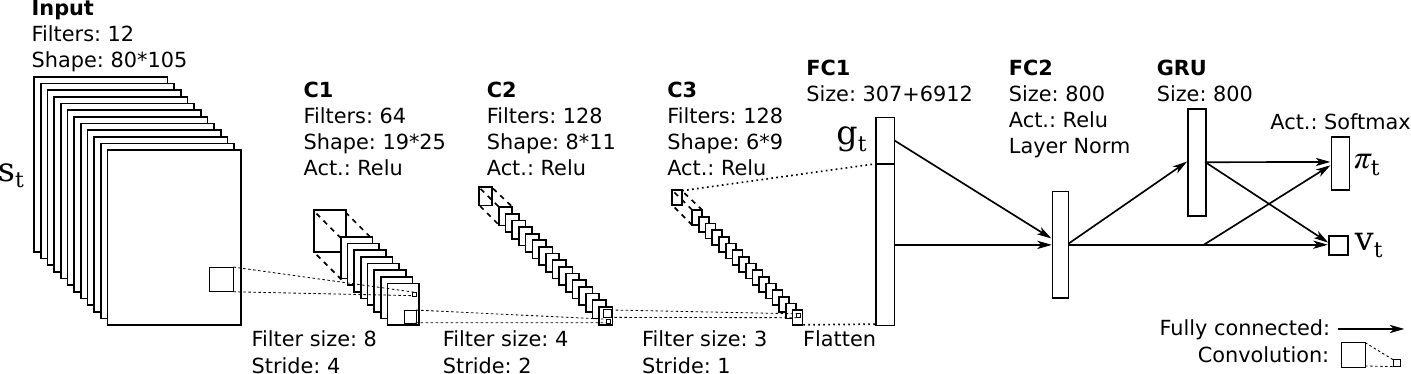}
            \caption{Policy-based Go-Explore architecture.}
    \end{subfigure}
    \caption{\textbf{Neural network architectures.} \textbf{(a)} The Atari architecture is based on the architecture provided with the backward algorithm implementation. The input consists of the RGB channels of the last four frames (re-scaled to 80 by 105 pixels) concatenated, resulting in 12 input channels. The network consists of 3 convolutional layers, 2 fully connected layers, and a layer of Gated Recurrent Units (GRUs)\cite{cho2014properties}. The network has a policy head $\pi(s|a)$ and a value head $V(s)$. \textbf{(b)} For the robotics problem, the architecture consists of two separate networks, each with 2 fully connected layers and a GRU layer. One network specifies the policy $\pi(s|a)$ by returning a mean $\mu$ and variance $\sigma$ for the actuator torques of the arm and the desired position of each of the two fingers of the gripper (gripper fingers are implemented as Mujoco position actuators\cite{todorov2012mujoco} with $kp=10^4$ and a control range of $[0, 0.05]$). The other network implements the value function $V(s)$. \textbf{(c)} The architecture for policy-based Go-Explore is identical to the Atari architecture, except that the goal representation $g$ is concatenated with the input of the first fully connected layer.}
    \label{efig:nn_architectures}
\end{figure}

\begin{figure}[ht!]
        \centering
        \begin{subfigure}[]{0.95\textwidth}
            \centering
            \includegraphics[width=\linewidth]{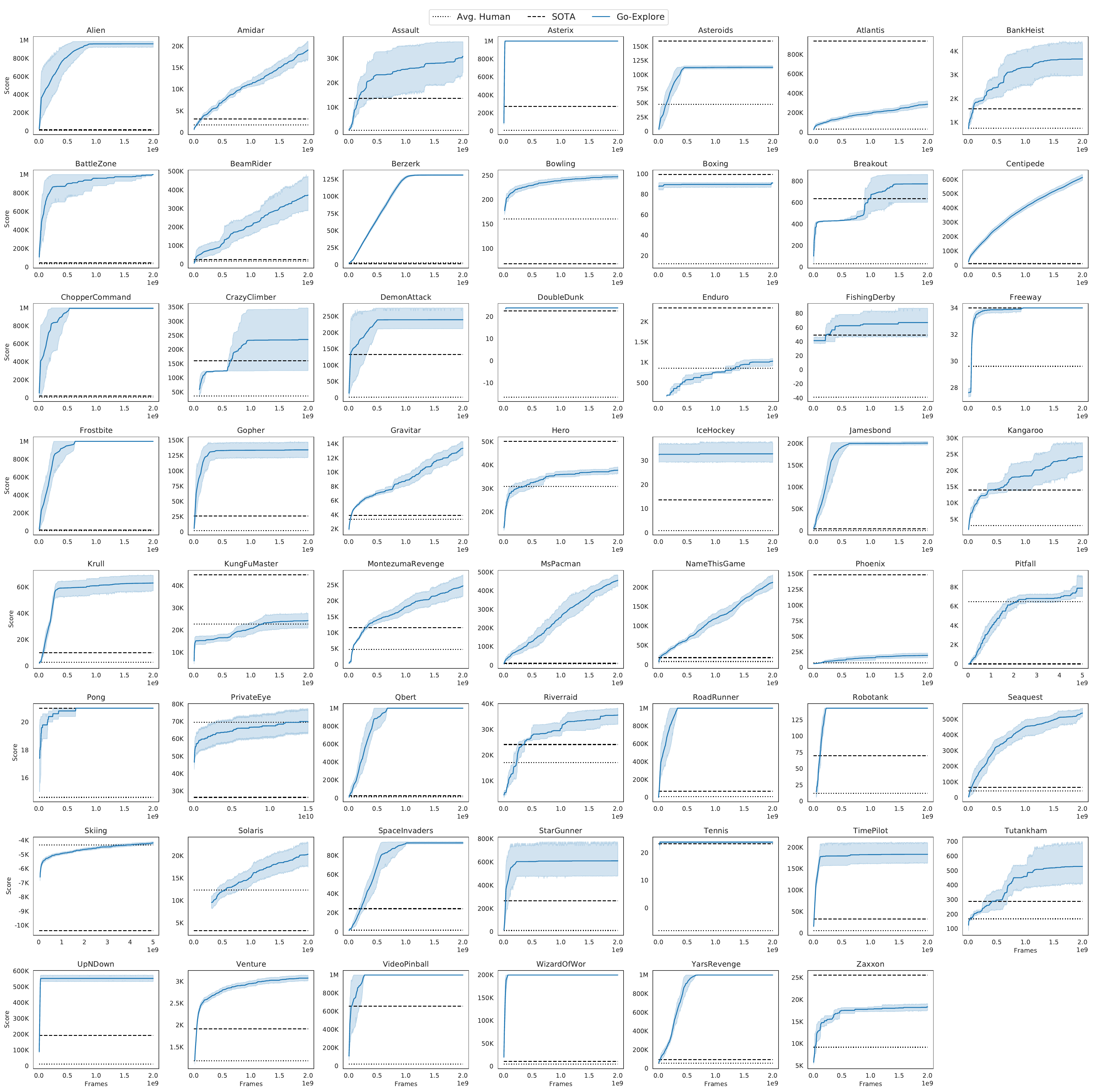}
            \caption{Exploration phase without domain knowledge.}
        \end{subfigure}
        \begin{subfigure}[]{\textwidth}
            \centering
            \includegraphics[width=0.3\linewidth]{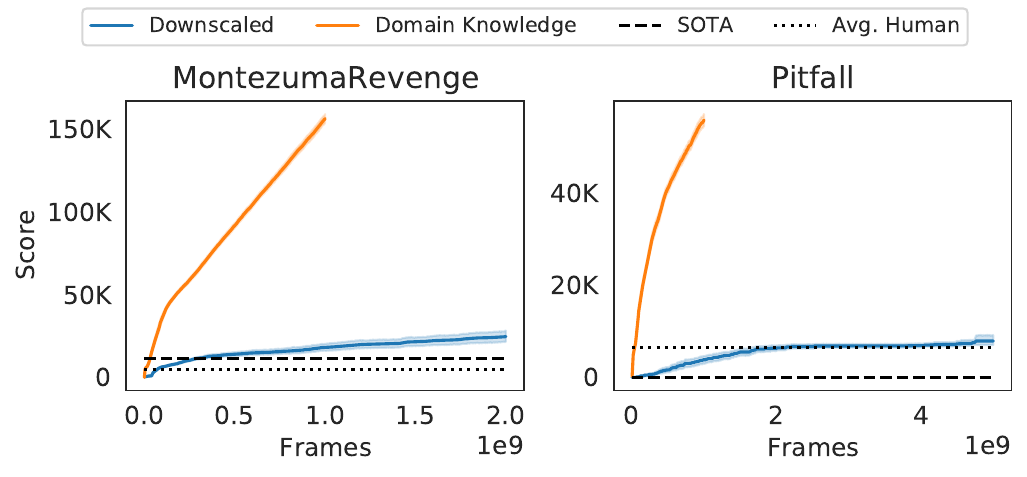}
            \caption{Exploration phase with domain knowledge (compared to downscaled)}.
        \end{subfigure}
        \caption{\textbf{Maximum end-of-episode score found by the exploration phase on Atari.} Because only scores achieved at the episode end are reported, the plots for some games (e.g.~Solaris) begin after the start of the run, when the episode end is first reached. In (a), averaging is over 50 runs for the 11 focus games and 5 runs for other games. In (b), averaging is over 100 runs.}
    \label{efig:phase1_atari_score}
\end{figure}

\begin{figure}[ht!]
        \centering
        \begin{subfigure}[]{0.95\textwidth}
            \centering
            \includegraphics[width=\linewidth]{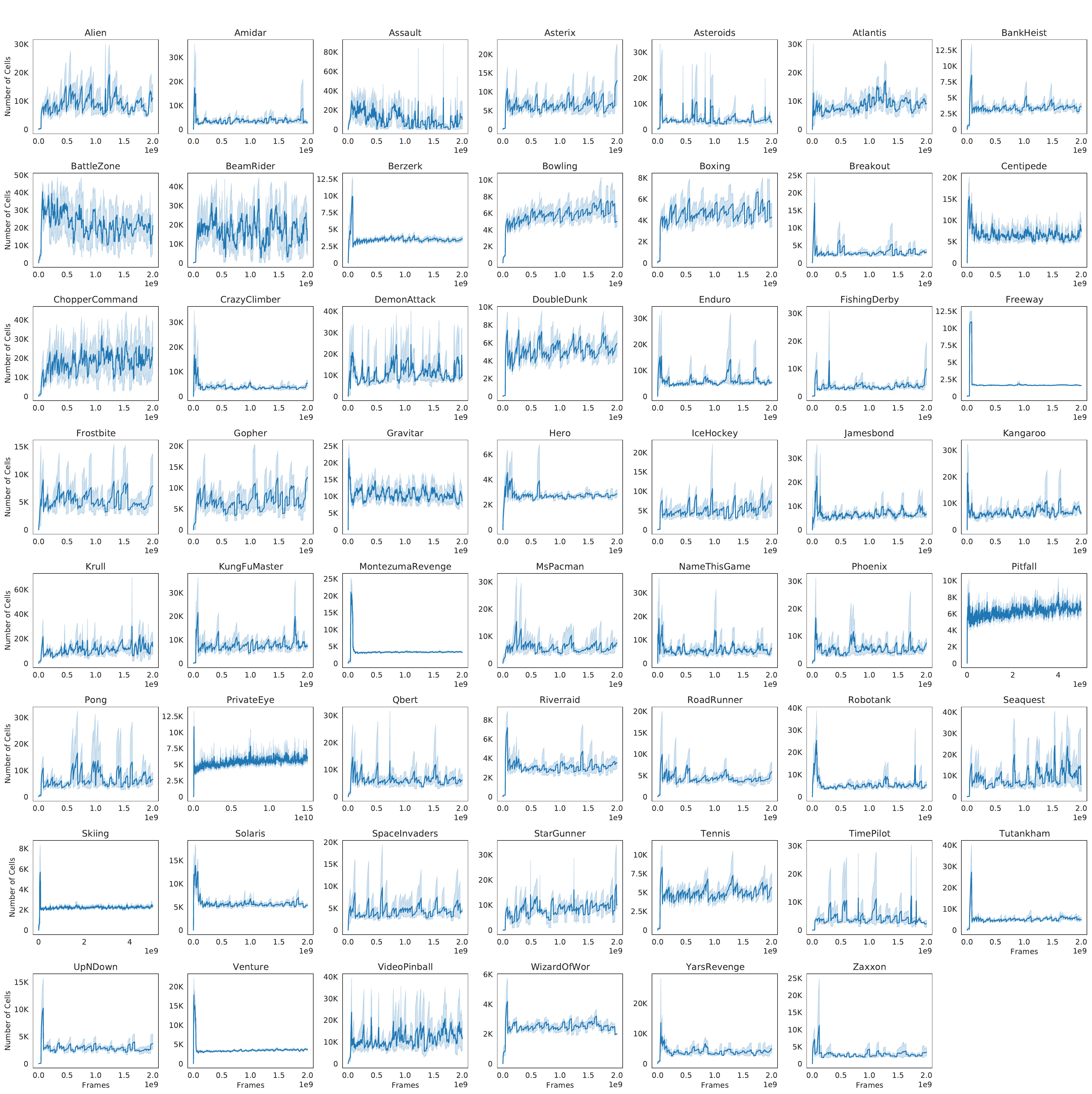}
            \caption{Exploration phase without domain knowledge.}
        \end{subfigure}
        \begin{subfigure}[]{\textwidth}
            \centering
            \includegraphics[width=0.3\linewidth]{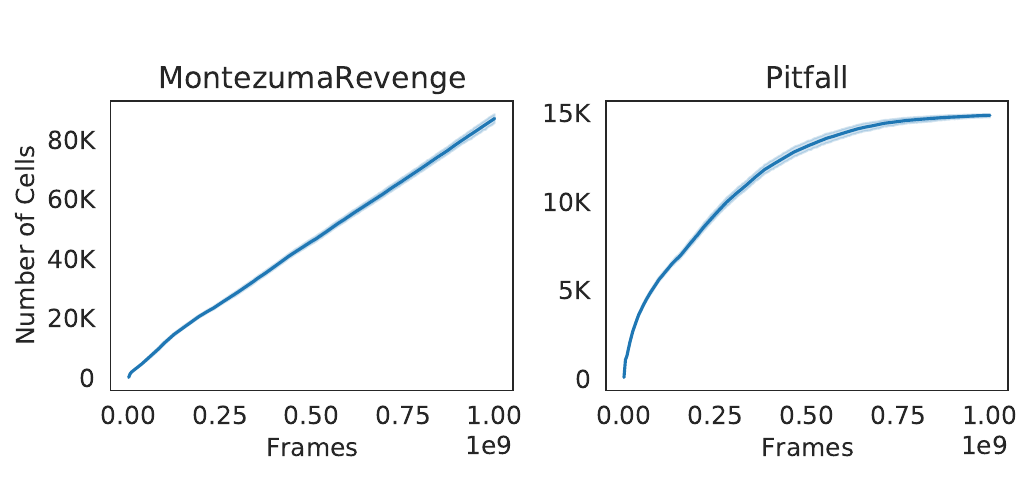}
            \caption{Exploration phase with domain knowledge.}
        \end{subfigure}
        \caption{\textbf{Number of cells in archive during the exploration phase on Atari.} In (a), archive size can decrease when the representation is recomputed. Previous archives are converted to the new format when the representation is recomputed, possibly leading to an archive larger than 50K. In this case, one iteration of the exploration phase runs and the representation is recomputed again.}
    \label{efig:phase1_atari_cells}
\end{figure}

\begin{figure}[ht!]
        \centering
        \begin{subfigure}[]{\textwidth}
            \centering
            \includegraphics[width=\linewidth]{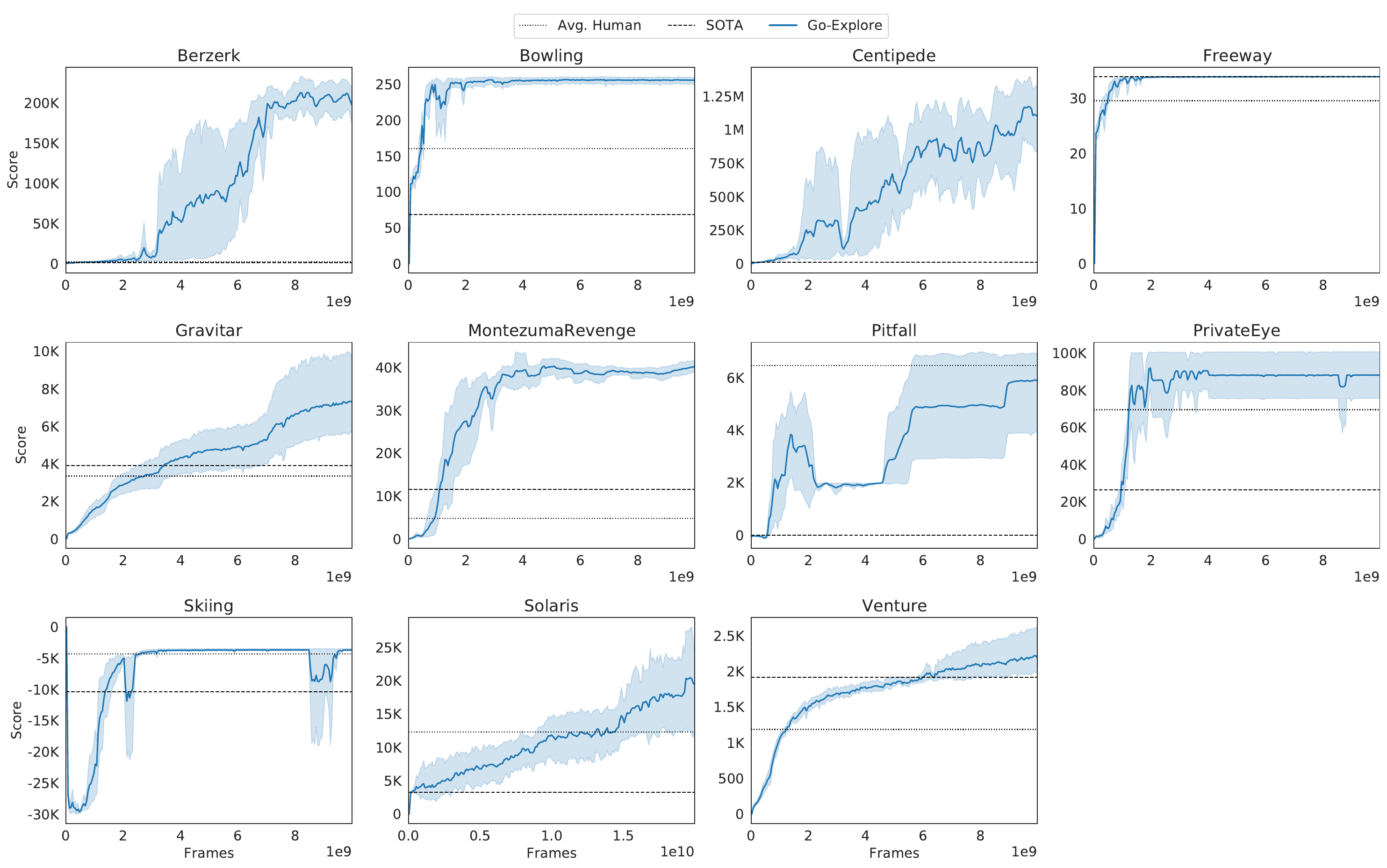}
            \caption{Exploration phase without domain knowledge.}
        \end{subfigure}
        \begin{subfigure}[]{\textwidth}
            \centering
            \includegraphics[width=0.55\linewidth]{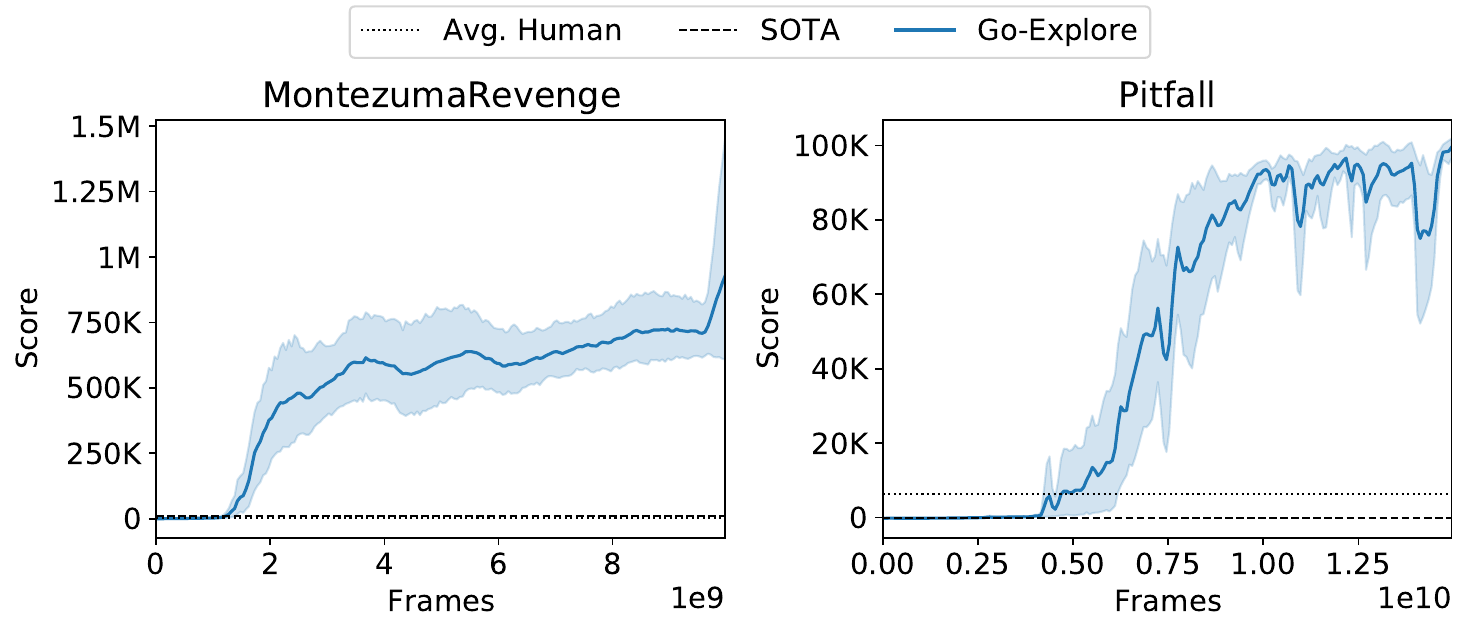}
            \caption{Exploration phase with domain knowledge.}
        \end{subfigure}
        \caption{\textbf{Progress of robustification phase on Atari.} Shown are the scores achieved by robustifying agents across training time for the exploration phase (a) with representations informed by domain knowledge, and (b) representations without domain knowledge. In particular, the rolling mean is shown of performance across the past 100 episodes when starting from the virtual demonstration (which corresponds to the domain's traditional starting state). Note that in (a) averaging is over 5 independent runs, while in (b) averaging is over 10 runs. Because the final performance is obtained by testing the highest-performing network checkpoint for each run over 1,000 additional episodes rather than directly extracted from the curves above, the performance reported in Figure~\ref{fig:phase2_sota} does not necessarily match any particular point along these curves (Methods).}
    \label{efig:phase2_atari}
\end{figure}

\begin{figure}[ht!]
    \centering
    \begin{subfigure}[t]{0.47\textwidth}
        \centering
        \includegraphics[width=\linewidth]{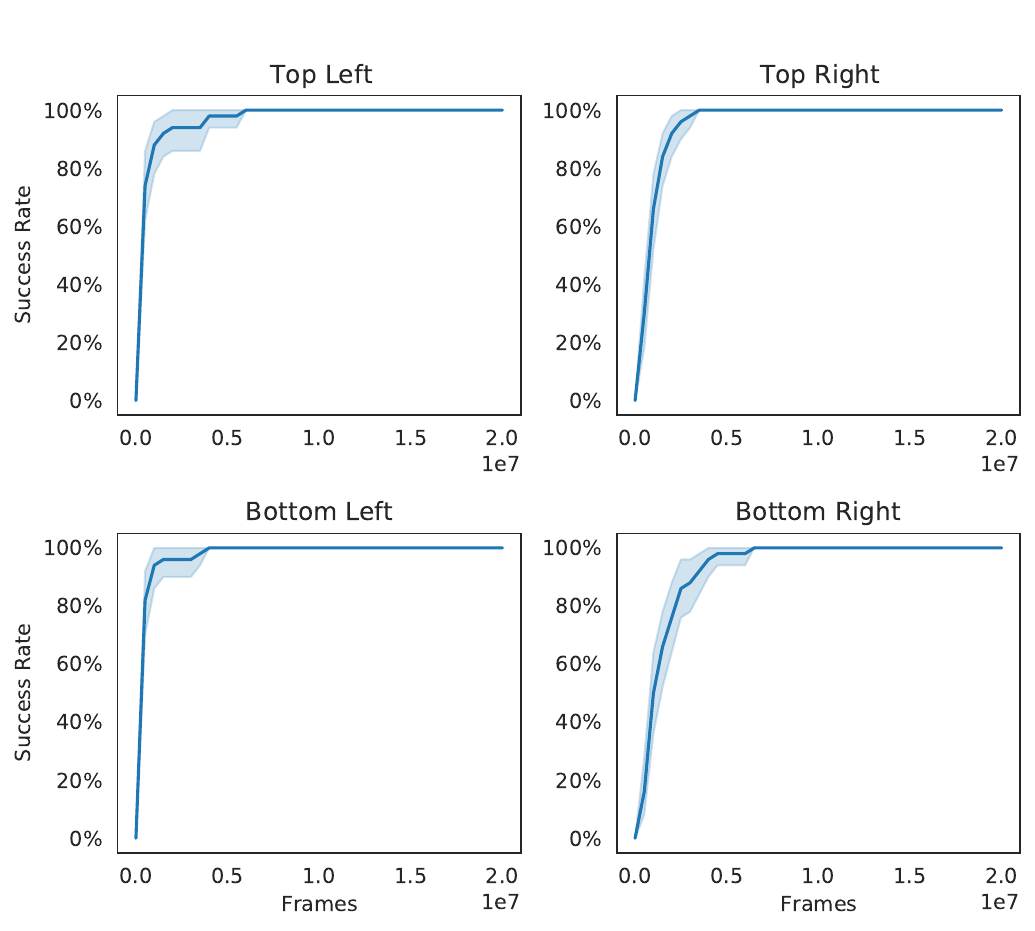}
        \caption{Runs with successful trajectories.}
        \label{efig:phase1_robotics_score}
    \end{subfigure}
    \begin{subfigure}[t]{0.47\textwidth}
        \centering
        \includegraphics[width=\linewidth]{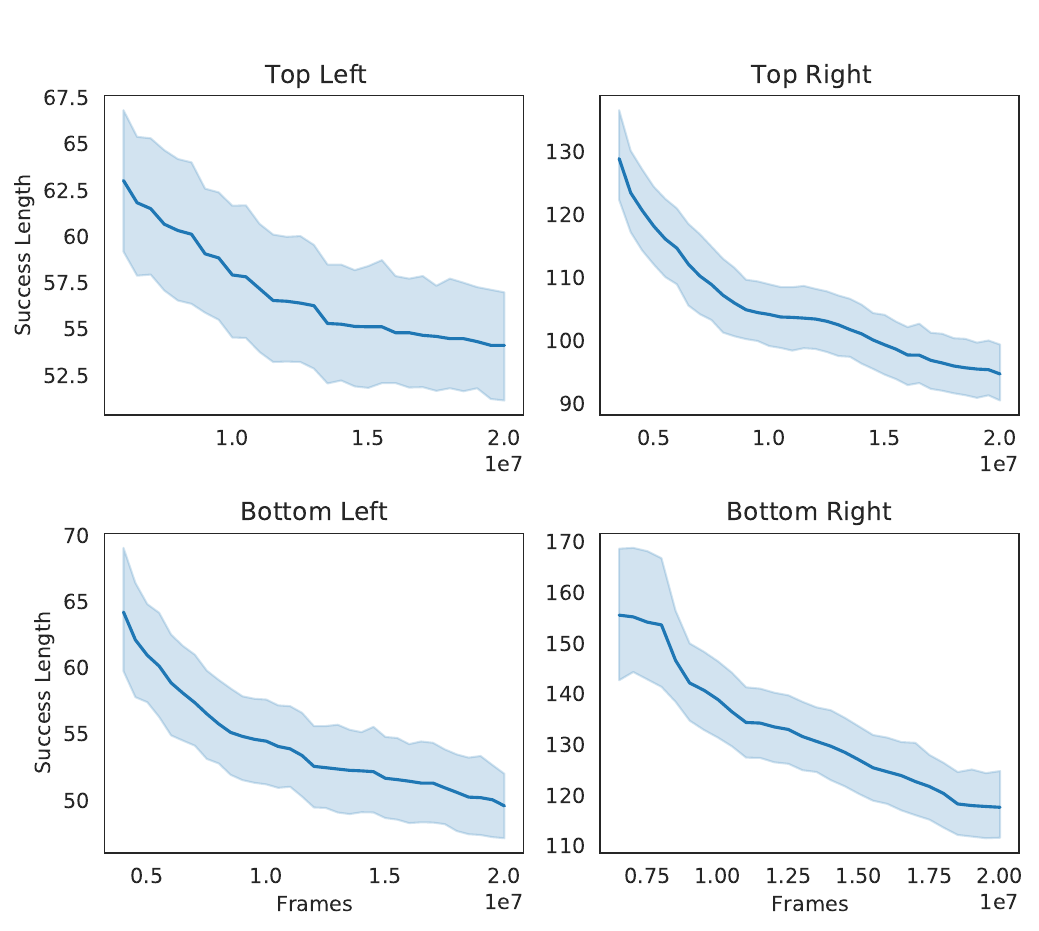}
        \caption{Length of the shortest successful trajectory}        \label{efig:phase1_robotics_length}
    \end{subfigure}
    \caption{\textbf{Progress of the exploration phase in the robotics environment.} In (a), the exploration phase quickly achieves 100\% success rate for all shelves in the robotics environment. However, (b) shows that while success is achieved quickly, it is useful to keep the exploration phase running longer to reduce the length of the successful trajectories, thus making robustification easier.}
\end{figure}

\begin{figure}[ht!]
        \centering
        \includegraphics[width=\linewidth]{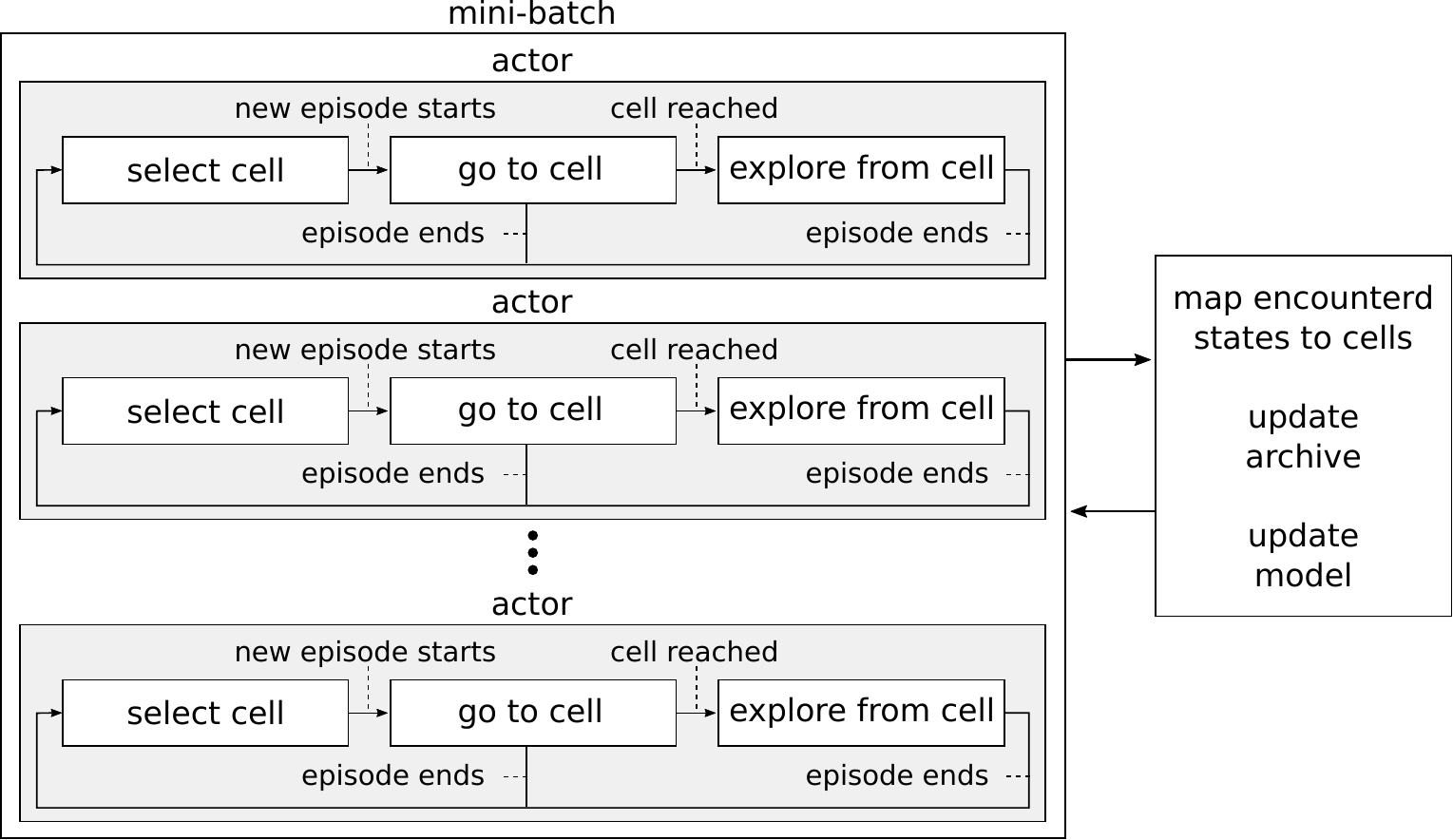}
        \caption{\textbf{Policy-based Go-Explore overview.} With respect to their practical implementation, the main difference between policy-based Go-Explore and Go-Explore without a policy is that in policy-based Go-Explore there exist separate actors that each have an internal loop switching between the ``select'', ``go'', and ``explore'' steps, rather than one outer loop in which the ``select'', ``go'', and ``explore'' steps are executed in synchronized batches. This structure allows policy-based Go-Explore to be easily combined with popular RL algorithms like A3C\cite{mnih2016asynchronous}, PPO\cite{Schulman2017ProximalPO} or DQN\cite{mnih:nature15}, which already divide data gathering over many actors.}
    \label{efig:policy_based_ge_overview}
\end{figure}

\begin{figure}[ht!]
    \centering
    \begin{subfigure}[t]{0.45\textwidth}
        \centering
        \includegraphics[width=\linewidth]{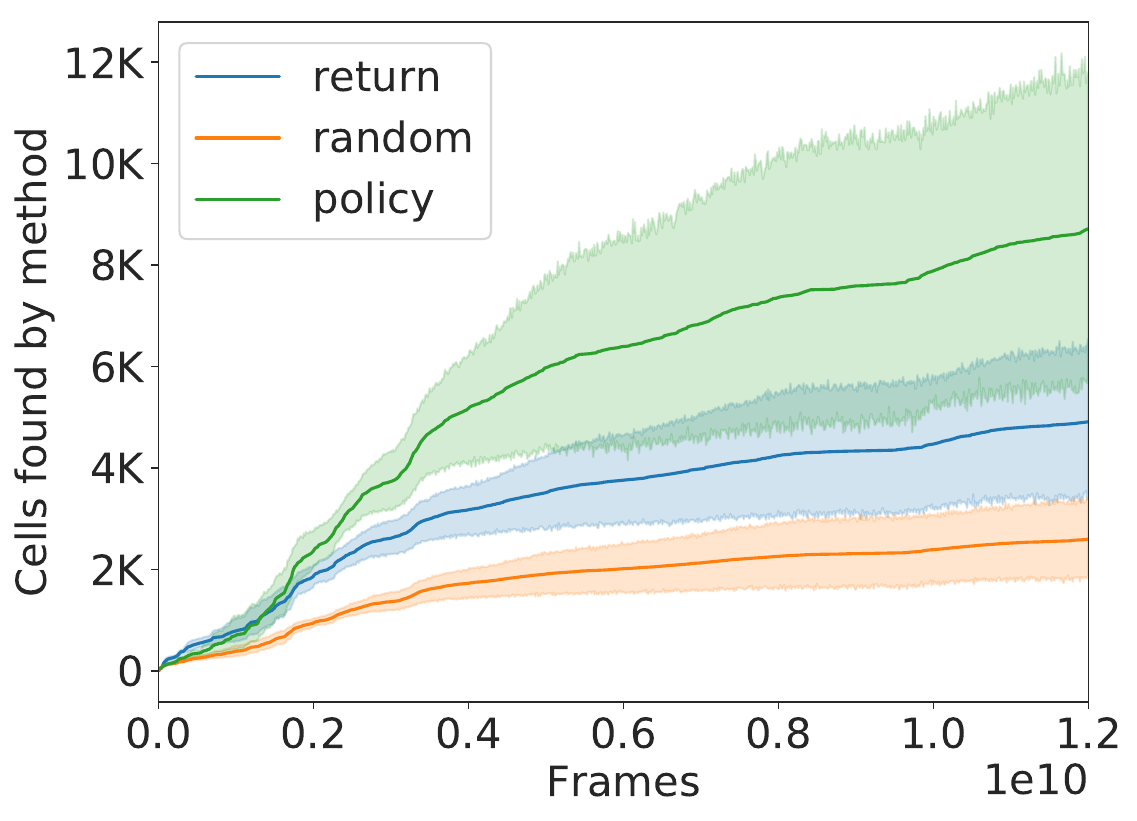}
        \caption{Montezuma's Revenge}
    \end{subfigure}%
    \begin{subfigure}[t]{0.45\textwidth}
        \centering
        \includegraphics[width=\linewidth]{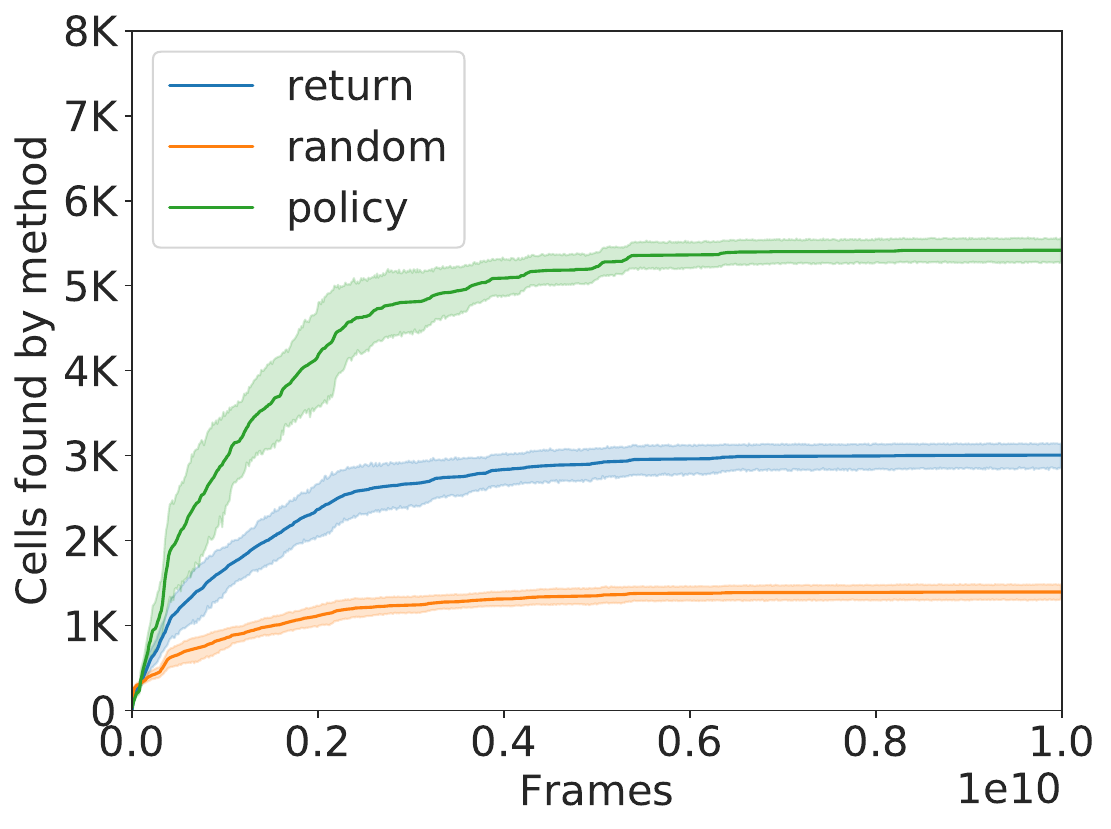}
        \caption{Pitfall}
    \end{subfigure}%
    \caption{\textbf{Method by which cells are found.} In both \textbf{(a)} Montezuma's Revenge and \textbf{(b)} Pitfall, sampling from the goal-conditioned policy results in the discovery of roughly four times more cells than when taking random actions. At the start of training, there is effectively no difference between random actions and sampling from the policy, supporting the intuition that sampling from the policy only becomes more efficient than random actions after the policy has acquired the basic skills for moving towards the indicated goal. Lastly, the number of cells that are discovered while returning is about twice that of the cells discovered when taking random actions after returning, indicating that the frames spent while returning to a previously visited cell are not just overhead required for moving towards the frontier of yet undiscovered states and training the policy network, but actually provide a substantial contribution towards exploration as well.}
    \label{efig:policy_based_ge_cell_discovery}
\end{figure}

\FloatBarrier

\begin{table}[ht!]
    \centering
    \fontsize{9}{10.5}\selectfont
    \begin{subfigure}[]{0.8\textwidth}
        \centering
        \begin{tabular}{c|cc|c}
            & \multicolumn{2}{c|}{\textbf{Robustification}} & \textbf{Policy-Based}\\ 
            \textbf{Parameter} & \emph{Atari} & \emph{Robotics} & \emph{Atari} \\
            \hline
            Discount factor ($\gamma$) & 0.999  & 0.99 & 0.99 \\
            N-step return factor ($\lambda$) & 0.95 & 0.95 & 0.95\\
            Nb. workers & 8 & 8 & 16\\
            Nb. actors per worker & 32 + 2 SIL & 120 + 8 SIL & 15 + 1 SIL\\
            Nb. actors ($N$) & 256 + 16 SIL & 960 + 64 SIL & 240 + 16 SIL\\
            Steps per batch ($T$) & 128 & 128 & 128 \\
            PPO Clip ($\epsilon$) & $0.1$ & $0.1$ & 0.1\\
            PPO Epochs & $4$ & $4$ & 4\\
            Value coef. ($w_{VF}$) & $0.5$ & 0.5 & 0.5\\
            Ent. coef. ($w_{ENT}$) & $10^{-5}$ & $10^{-5}$ & $10^{-4}$\\
            L2 coef. ($w_{L2}$) & $10^{-7}$ & $10^{-7}$ & $10^{-7}$\\
            SIL coef. ($w_{SIL}$) & 0.1 & 0.1 & 0.1 \\
            SIL ent. coef. ($w_{SIL\_ENT}$) & $10^{-5}$ & $10^{-5}$ & 0\\
            SIL value coef. ($w_{SIL\_VF}$) & 0.01 & 0.1 & 0.01\\
            Cell selection weight & $W = \dfrac{1}{\sqrt{C_\mathrm{seen} + 1}}$ & $W = \dfrac{1}{\sqrt{C_\mathrm{seen} + 1}}$ & $W=\dfrac{1}{0.5 C_\mathrm{steps}+1}$\\
            $x,y$ coordinates discretization & 8 by 16 pixels & - & 18 by 18 pixels \\
            \hline
            Allowed lag & 50 & 10 & - \\
            Extra frame coef & 7 & 4 & -\\
            Move threshold & 0.1 & 0.1 & -\\
            Nb. demonstrations & 10 + 1 virtual & 10 & -\\
            SIL from start prob. & 0.3 & 0 & -\\
            Window size (frames) & 160 & 40 & -\\
        \end{tabular}
        \caption{Robustification and policy-based parameters}
        \label{etab:backward_hyper_atari}
    \end{subfigure}
    \begin{subfigure}[]{0.3\textwidth}
        \centering
        \renewcommand{\footnotesize}{\fontsize{8}{10}\selectfont}
        \begin{minipage}{\linewidth}
            \centering
            \vspace{4mm}
            \begin{tabular}{c|c}
                \textbf{Parameter} & \textbf{Value} \\
                \hline
                Sticky actions & True \\
                Length limit & 400K frames\footnote{OpenAI Gym default} \\
                End of episode & All lives lost\footnote{Except Montezuma's Revenge with return policy (Methods)} \\
                Action repeat & 4 \\
                Frame max pool & 2 or 4\footnote{4 for Gravitar and Venture}\\
            \end{tabular}
        \end{minipage}
        \caption{Atari environment parameters}
        \label{etab:atari_hyper}
    \end{subfigure}
    \caption{\textbf{Hyperparameters.} (a) Parameters above the dividing line are applicable to PPO in general, while parameters below the line are specific to the backward algorithm. ``Allowed lag'' is the number of frames the agent may lag the demonstration before being considered unsuccessful. When the agent matches the demonstration, it runs for additional frames, controlled by ``Extra frame coef'' $c$: $\lfloor e^{cX} \rfloor$ ($X \sim U(0, 1)$). Window size is the number of starting points below the maximum starting point of the demonstration that the algorithm may start from. In the equations specifying cell selection weight, $C_\mathrm{seen}$ is the number of exploration steps in which that cell is visited (i.e. the $C_\mathrm{seen}$ count of a cell is increased by one when it is visited in the exploration step, even if the cell was visited multiple times in that step). $C_\mathrm{steps}$ is the total number of steps the agent spend in the cell. The $x,y$ discretization only refers to experiments with domain knowledge and the top 50 pixels are ignored as the agent can never enter them. (b) For the exploration phase, only ``Max episode length'', ``End of episode'', and ``Action repeat'' apply.}
    \label{etab:hyper}
\end{table}

\begin{table}[htbp]

    \centering
    \fontsize{9}{10.5}\selectfont
    \begin{subfigure}[]{0.45\textwidth}
        \centering
        \begin{tabular}{c}
            \textbf{Object} \\
            \hline
            \texttt{door1} \\
            \texttt{door} \\
            \texttt{elbow\_flex\_link} \\
            \texttt{forearm\_roll\_link} \\
            \texttt{gripper\_link} \\
            \texttt{head\_camera\_link} \\
            \texttt{head\_pan\_link} \\
            \texttt{head\_tilt\_link} \\
            \texttt{l\_gripper\_finger\_link} \\
            \texttt{latch1} \\
            \texttt{latch} \\
            \texttt{obj0} \\
            \texttt{r\_gripper\_finger\_link} \\
            \texttt{shoulder\_lift\_link} \\
            \texttt{shoulder\_pan\_link} \\
            \texttt{upperarm\_roll\_link} \\
            \texttt{wrist\_flex\_link} \\
            \texttt{wrist\_roll\_link}
        \end{tabular}
        \caption{Position and velocity objects}
        \label{etab:robotics_pos}
    \end{subfigure}
    \quad \quad
    \begin{subfigure}[]{0.45\textwidth}
        \centering
        \begin{tabular}{c}
            \textbf{Object} \\
            \hline
                \texttt{DoorLR} \\
                \texttt{DoorUR} \\
                \texttt{Shelf} \\
                \texttt{Table} \\
                \texttt{door1} \\
                \texttt{door} \\
                \texttt{frameL1} \\
                \texttt{frameL} \\
                \texttt{frameR1} \\
                \texttt{frameR} \\
                \texttt{gripper\_link} \\
                \texttt{l\_gripper\_finger\_link} \\
                \texttt{latch1} \\
                \texttt{latch} \\
                \texttt{obj0} \\
                \texttt{r\_gripper\_finger\_link} \\
                \texttt{world}
        \end{tabular}
        \caption{Collision and bounding box objects}
        \label{etab:robotics_interest}
    \end{subfigure}
    \caption{\textbf{Robotics state representation.} Position and velocities of the objects in (a) are included in the state representation for robotics. Collisions between any two objects in (b) as well as whether each object is currently inside the bounding boxes for the table and shelves are also included in the state representation. Objects are given by their MuJoCo\cite{todorov2012mujoco} entity names in the source code for the environment. Door-related objects ending with a 1 correspond to the lower door while door-related objects not ending with anything correspond to the upper door. The \texttt{frame} objects are the unmovable wooden blocks situated on either side of the movable part of the door. ``L'' and ``R'' correspond to ``left'' and ``right'' whereas ``L'' and ``U'' correspond to ``lower'' and ``upper''. The difference between ``door'' and ``DoorUR'' as well as ``door1'' and ``DoorLR'' is that in each case the latter object corresponds to the entire door structure, including the frames, while the former corresponds only to the movable part of the door. A link to the original source code for the MuJoCo description files defining these entities is given in ``Acknowledgements'', and a link to the Go-Explore codebase containing our modified version is provided in Methods section ``Code availability''.}
\end{table}

\begin{table}[ht!]
    \centering
    \fontsize{9}{10.5}\selectfont
    \begin{tabular}{l|r|rrr}
    \bf{Game} &   \bf{Expl. Phase} &     \bf{Robust. Phase} &      \bf{SOTA} & \bf{Avg. Human}\\
    \hline
Alien            & 959,312 &             &  11,358 &     7,128 \\
Amidar           &  19,083 &             &   3,092 &     1,720 \\
Assault          &  30,773 &             &  13,759 &       742 \\
Asterix          & 999,500 &             & 274,491 &     8,503 \\
Asteroids        & 112,952 &             & 159,426 &    47,389 \\
Atlantis         & 286,460 &             & 937,558 &    29,028 \\
BankHeist        &   3,668 &             &   1,563 &       753 \\
BattleZone       & 998,800 &             &  45,610 &    37,188 \\
BeamRider        & 371,723 &             &  24,031 &    16,927 \\
Berzerk          & 131,417 &     \bf{197,376} &   1,383 &     2,630 \\
Bowling          &     247 &       \bf{260} &      69 &       161 \\
Boxing           &      91 &             &      99 &        12 \\
Breakout         &     774 &             &     637 &        31 \\
Centipede        & 613,815 & \bf{1,422,628} &  10,166 &    12,017 \\
ChopperCommand   & 996,220 &             &  19,256 &     7,388 \\
CrazyClimber     & 235,600 &             & 160,161 &    35,829 \\
DemonAttack      & 239,895 &             & 133,030 &     1,971 \\
DoubleDunk       &      24 &             &      23 &       -16 \\
Enduro           &   1,031 &             &   2,338 &       861 \\
FishingDerby     &      67 &             &      49 &       -39 \\
Freeway          &      34 &        \bf{34} &      \bf{34} &        30 \\
Frostbite        & 999,990 &             &  10,003 &     4,335 \\
Gopher           & 134,244 &             &  26,123 &     2,413 \\
Gravitar         &  13,385 &     \bf{7,588} &   3,906 &     3,351 \\
Hero             &  37,783 &             &  50,142 &    30,826 \\
IceHockey        &      33 &             &      14 &         1 \\
Jamesbond        & 200,810 &             &   4,303 &       303 \\
Kangaroo         &  24,300 &             &  13,982 &     3,035 \\
Krull            &  63,149 &             &   9,971 &     2,666 \\
KungFuMaster     &  24,320 &             &  44,920 &    22,736 \\
MontezumaRevenge &  24,758 &    \bf{43,791} &  11,618 &     4,753 \\
MsPacman         & 456,123 &             &   9,901 &     6,952 \\
NameThisGame     & 212,824 &             &  18,084 &     8,049 \\
Phoenix          &  19,200 &             & 148,840 &     7,243 \\
Pitfall          &   7,875 &     \bf{6,954} &       0 &     6,464 \\
Pong             &      21 &             &      21 &        15 \\
PrivateEye       &  69,976 &    \bf{95,756} &  26,364 &    69,571 \\
Qbert            & 999,975 &             &  26,172 &    13,455 \\
Riverraid        &  35,588 &             &  24,116 &    17,118 \\
RoadRunner       & 999,900 &             &  67,962 &     7,845 \\
Robotank         &     143 &             &      70 &        12 \\
Seaquest         & 539,456 &             &  64,985 &    42,055 \\
Skiing           &  -4,185 &    \bf{-3,660} & -10,386 &    -4,337 \\
Solaris          &  20,306 &    \bf{19,671} &   3,282 &    12,327 \\
SpaceInvaders    &  93,147 &             &  24,183 &     1,669 \\
StarGunner       & 609,580 &             & 265,480 &    10,250 \\
Tennis           &      24 &             &      23 &        -8 \\
TimePilot        & 183,620 &             &  32,813 &     5,229 \\
Tutankham        &     528 &             &     288 &       168 \\
UpNDown          & 553,718 &             & 193,520 &    11,693 \\
Venture          &   3,074 &     \bf{2,281} &   1,916 &     1,188 \\
VideoPinball     & 999,999 &             & 656,572 &    17,668 \\
WizardOfWor      & 199,900 &             &  10,980 &     4,757 \\
YarsRevenge      & 999,998 &             &  93,680 &    54,577 \\
Zaxxon           &  18,340 &             &  25,603 &     9,173 \\
    \end{tabular}
    \caption{\textbf{Full scores on Atari.}}
    \label{etab:atari_full_scores}
\end{table}

\FloatBarrier

\endrefsegment
\newrefsegment

\section*{Supplementary Information}

\renewcommand{\figurename}{Supplementary Information Figure}
\renewcommand{\tablename}{Supplementary Information Table}
\setcounter{figure}{0}
\setcounter{table}{0}

\subsection{Detachment and derailment in reinforcement learning}

Environment exploration has long been a central topic in the field of reinforcement learning\cite{bellemare2016unifying, guo2019efficient, burda:rnd2018, Choi2018ContingencyAwareEI, tang2017exploration}. 
Despite the extensive research on exploration in reinforcement learning, we hypothesise that many previous algorithms have been affected by two major issues which we call \emph{detachment} and \emph{derailment}.

\begin{figure}[tb!]
    \centering
    \includegraphics[width=\linewidth]{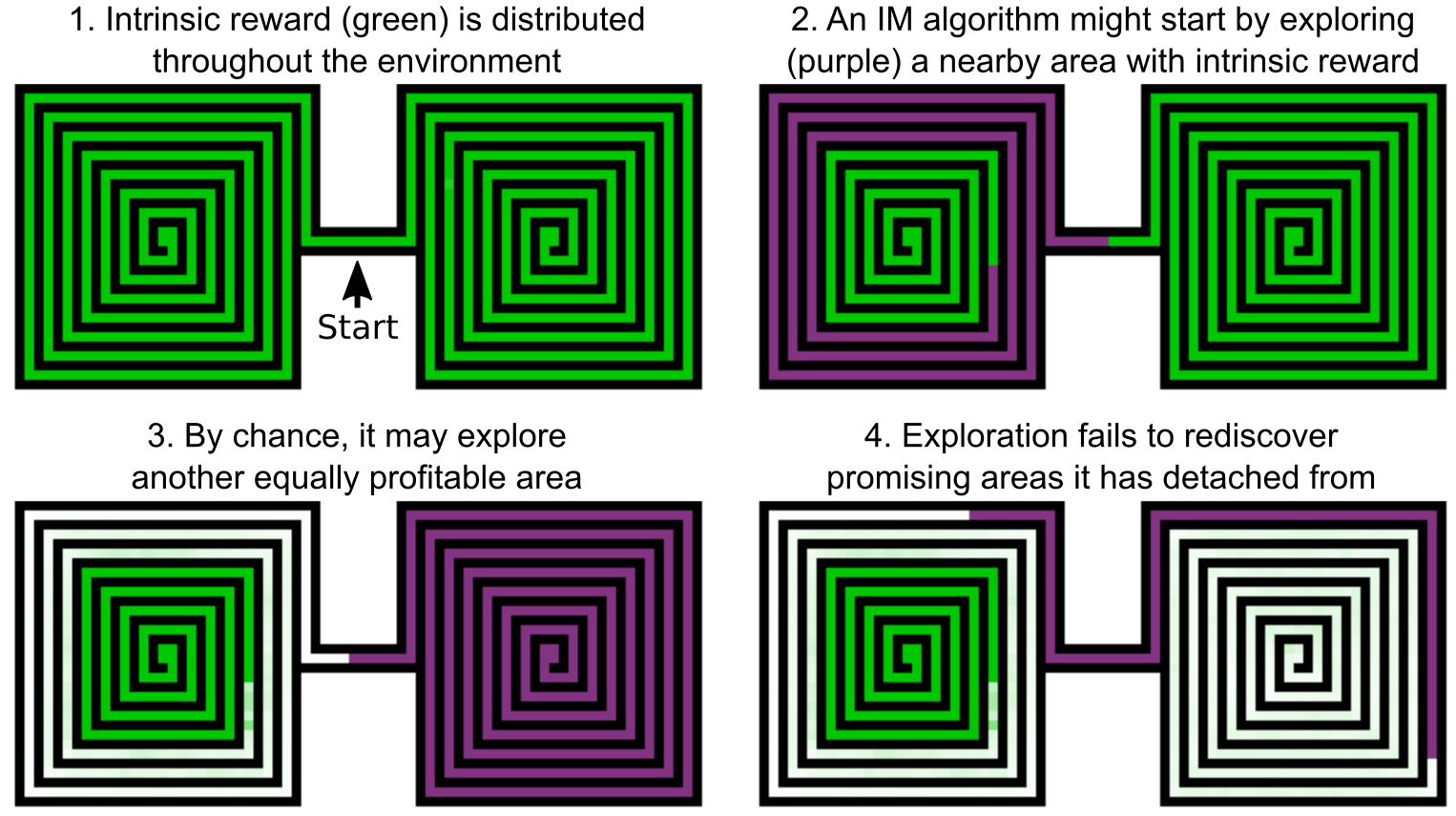}
    \caption{\textbf{Example of detachment with intrinsic reward.} Green areas indicate intrinsic reward, white indicates areas where no intrinsic reward remains, and purple areas indicate where the algorithm is currently exploring.}
    \label{fig:detachment}
\end{figure}

We define detachment as losing track of interesting areas to explore from. 
Here, ``interesting areas to explore from'' refers to states for which we have evidence (e.g. a low number of visits) that they could lead to the discovery of new areas of the environment.
``Losing track'' means that the algorithm stops trying to visit those areas prematurely, despite the fact that they are not thoroughly explored yet.
For reinforcement learning algorithms that only optimise the expected return, detachment is almost guaranteed, as these algorithms do not attempt to promote exploration explicitly. Unless external rewards are aligned with interesting areas to explore from, such algorithms will stop visiting under-explored areas in favour of areas with a high return.
However, algorithms that reward the agent for exploring new states can still suffer from detachment.
In intrinsic-motivation (IM) algorithms, for example,  detachment can happen because the intrinsic reward is lowered each time a state is visited, so that, eventually, they in effect provide no incentive for an agent to return to them.
This issue can be especially prominent when there are multiple frontiers, as the agent may now, due to stochastic exploration, stop visiting one of those frontiers long enough to forget how to return to it (Fig.~\ref{fig:detachment}). 
When this happens, the agent has to effectively relearn how to reach the frontier from scratch, but this time there is no intrinsic reward anymore to guide the agent.
If intrinsic rewards were required to find the frontier in the first place, meaning that an algorithm without intrinsic rewards would fail to find this frontier, the IM algorithm will similarly fail to rediscover the frontier as well, meaning it has detached from the frontier. 
One might think that allowing intrinsic motivation to regrow after a time would solve the issue, but in that case the same dynamic can just play out over and over again endlessly.

We define derailment as when the exploratory mechanisms of the algorithm prevent it from returning to previously visited states.
Returning to previously visited states is important because many RL environments, and especially hard-exploration environments, contain a large number of states that are far way from a starting state and can not be easily reached from such a starting state through random actions, even when exploring for hundreds of billions of frames.
To discover these states, many algorithms rely on the policy learning to take actions that lead to states that are increasingly further away, either because the policy discovered external rewards leading towards these far away states, or because the algorithm provides intrinsic rewards that lead the policy towards infrequently visited states.
However, as the policy needs to take an increasingly large number of correct actions to reach unexplored areas of the environment, it becomes increasingly likely that a state-agnostic exploration mechanism (i.e. any exploration mechanism that explores the same amount in all states, regardless of whether the agent is in a novel or a well-explored area), such as $\epsilon$-greedy exploration, will cause the policy to take one or more exploratory actions that prevent it from reaching the distant state it sought to return to, thus stifling exploration.
Two common strategies to prevent derailment are to (1) set the exploration probability (e.g. $\epsilon$ in $\epsilon$-greedy exploration) to be small or (2) to start with a high exploration probability, but reduce it over training iterations\cite{sutton1998reinforcement}.
Working with a fixed, low exploration probability throughout training reduces the effect derailment throughout training, but it also means that very little exploration will happen once a new state is reached.
Annealing exploration over training iterations will initially lead to a lot of exploration at the cost of heavy derailment. 
Unfortunately, if there exists a far away state that requires precise actions to reach, that state will not be reached until the exploration probability has been reduced sufficiently to avoid derailment.
This low exploration probability means that, once the agent is finally able reliably reach this far away state, very little exploration will be performed after reaching it.

The solution we propose to avoid derailment is to have an algorithm exhibit separate exploration probabilities depending on whether it is in a well-known area of the environment, meaning the probability of exploratory actions should be low, or in an unknown area of the environment, meaning the probabilities of exploratory actions should be high.
While doing so is not a feature of state-agnostic exploratory mechanisms like $\epsilon$-greedy exploration, one could hope that it is a property of algorithms that explore by sampling from a stochastic policy, because stochastic policies could learn to be low entropy in familiar states (i.e. they are certain about the correct action) while remaining high entropy in new states (i.e. where they should be uncertain about the correct action).
However, deep learning struggles to remain well-calibrated when provided out-of-distribution input data. 
In image classification, for example, when networks classify images far out of distribution, it would be helpful if these networks returned a uniform distribution of probability across classes to properly indicate uncertainty, but networks instead are often surprisingly overconfident\cite{nguyen2015deep, szegedy:corr13}.
In the context of RL, this result means that we can expect trained policies to be highly confident in their actions (i.e.\ low entropy), even in areas that they have never observed before, thus resulting in a lack of exploration.
To remedy this issue, stochastic-policy RL algorithms generally add an entropy bonus to the loss function, encouraging the policy to assign more equal probability to all actions.
However, because this entropy bonus applies equally throughout the trajectory of a policy, it is difficult to tune the entropy bonus in a way that guarantees effective exploration without sacrificing the network's ability to return; if the entropy bonus is too high, the policy will frequently take exploratory actions that prevent it from returning, thus causing derailment, but if the entropy bonus is too small, the policy will not explore sufficiently when a new area is reached.

\subsection{Derailment in robotics}

As shown in the main text of this paper, a count-based intrinsic motivation control completely fails to discover any rewards in the robotics environment even though it is given the same domain knowledge state representation as Go-Explore's exploration phase and when given a comparable budget of frames to Go-Explore's exploration and robustification phases combined. 
Evidence from the experiments suggests that this failure is primarily due to the problem of \emph{derailment}, specifically to the difficulty that the IM control has of learning to reliably \emph{grasp} the object.

Grasping is widely considered an extremely difficult task to learn in robotics\cite{nair2018overcoming,kraft2010development}. The overwhelming majority of undiscovered cells are those require grasping the object and lifting it to reach.
The claim that the failure to explore the environment is due to derailment when grasping the object necessitates that grasping is \emph{discovered}, but cannot be reliably reproduced by the policy due to its excessive exploratory mechanisms. We separate the discovery of grasping into three steps: touching the object with one of the two grippers (the ``touch'' step), touching that surrounds the object with both grippers (the ``grasp'' step), and finally lifting the object (the ``lift'' step). An analysis of the cells and counts discovered by 20 control runs (5 per target shelf) shows that all runs discover the ``touch'' and ``grasp'' step, but in 18 (90\%) of these runs, the count associated with the ``grasp'' step is at least 10x smaller than that associated with the ``touch'' step, indicating difficulty (and thus possible derailment) in learning to go from the ``touch'' step to the ``grasp'' step. In the 2 (10\%) remaining runs, the ``grasp'' step count is closer to the ``touch'' step count, but, in one case, lifting is never discovered, and in the other, lifting is discovered, but the count for the ``lift'' step is again over 10x smaller than that of the ``grasp'' step, indicating possible derailment in between those two steps. It is thus apparent that the IM control has difficulty returning to the grasping stepping stones that it discovers, in spite of these cells often having amongst the lowest counts of any cell discovered, and thus the highest intrinsic rewards, thereby providing evidence of derailment.

\subsection{Policy-based Go-Explore and Stochastic Environments}

Restoring simulator state is a highly efficient method for returning to previously visited states, as it both removes the need to replay the trajectory towards a previously visited state as well as the need to train a policy capable of doing so reliably. 
That said, doing so also has the potential drawback that some of the trajectories found by restoring simulator state can be hard to robustify if that trajectory is not representative of realistic policies that can succeed in the stochastic testing environment.
For example, imagine a robot with the goal of crossing a busy highway.
The cars on the highway are stochastic, meaning that their position, speed, and reactions will differ in every episode, but the highway is always busy.
As such, we assume that there is no safe way to reliably cross the busy highway directly.
The highway has an overpass that allows the robot to easily and reliably cross the highway safely, but it is located some distance away from the robot, meaning that it is not the shortest method to cross the highway. 
However, when restoring simulator state, it is likely that Go-Explore will find a way directly across the highway in this particular scenario, because there probably exists some static sequence of lucky actions that brings the agent from one cell on the highway to the next cell, and once the next cell is reached, that progress is saved in the form of the simulator state. 
Once the opposite side of the highway is reached, this shorter trajectory will overwrite any longer trajectories that go over the overpass, and the final trajectory returned will go directly over the highway.
Training a policy that can reliably follow this trajectory may be impossible because the inherent stochasticity of the cars on the highway (i.e. the inherent stochasticity of the environment) can make it such that a sufficiently reliable policy simply does not exist. That is, each new random busy highway situation requires its own lucky set of actions that may not derive in any systematic way from the agent's observations.

Policy-based Go-Explore can alleviate this situations in two ways.
First, because its progress along the highway is not saved and because each trial is in a stochastic environment, policy-based Go-Explore must attempt to return manually to each cell across the highway in different conditions. 
If there does not exist a reliable policy that can do so, it is unlikely that policy-based Go-Explore will ever cross the highway this way.
As a result, policy-based Go-Explore is much more likely to learn a policy that reliably navigates the overpass instead.

Second, because policy-based Go-Explore needs to return to cells in the presence of stochasticity, it can keep track of the success rate towards each cell in the archive. 
As such, even if policy-based Go-Explore is sometimes able to cross the highway, it is possible to not overwrite cells on the other side of the highway until the policy has learned to return to those cells reliably.
Doing so prevents the shorter, but unreliable trajectories from overwriting the longer but more reliable trajectories that take the safe overpass.
A similar mechanic could be implemented to deal with stochasticity in rewards, as policy-based Go-Explore makes it possible to track the average reward when attempting to reach a particular state.
Because it was not necessary to implement such mechanics for the games of Montezuma's Revenge and Pitfall, studying the effectiveness and exact implementation details of such a mechanic is a topic for future research.

Last, it may be possible to resolve these issues even when Go-Explore is allowed to restore simulator state. 
For example, it is possible to run the Go-Explore exploration phase many times with different random seeds, thus making it possible to estimate which trajectories are and which trajectories are not reliable. Recognising that trajectories with only slight differences in the states visited still represent effectively the same solution could be handled by the aggregation that occurs in the cell representation, meaning that solutions that visit the same \emph{cells} in order (instead of the same states in order) could be considered the same.
It thus seems possible to produce a version of Go-Explore that is able to estimate the reliability of different trajectories while still gaining the advantages of restoring simulator state. Identifying the exact form of that algorithm and experimentally validating it is a fruitful area of future research.

\subsection{Policy-based Go-Explore vs. DTSIL}

\begin{figure}
    \centering
    \begin{subfigure}[t]{0.45\textwidth}
        \centering
        \includegraphics[width=\linewidth]{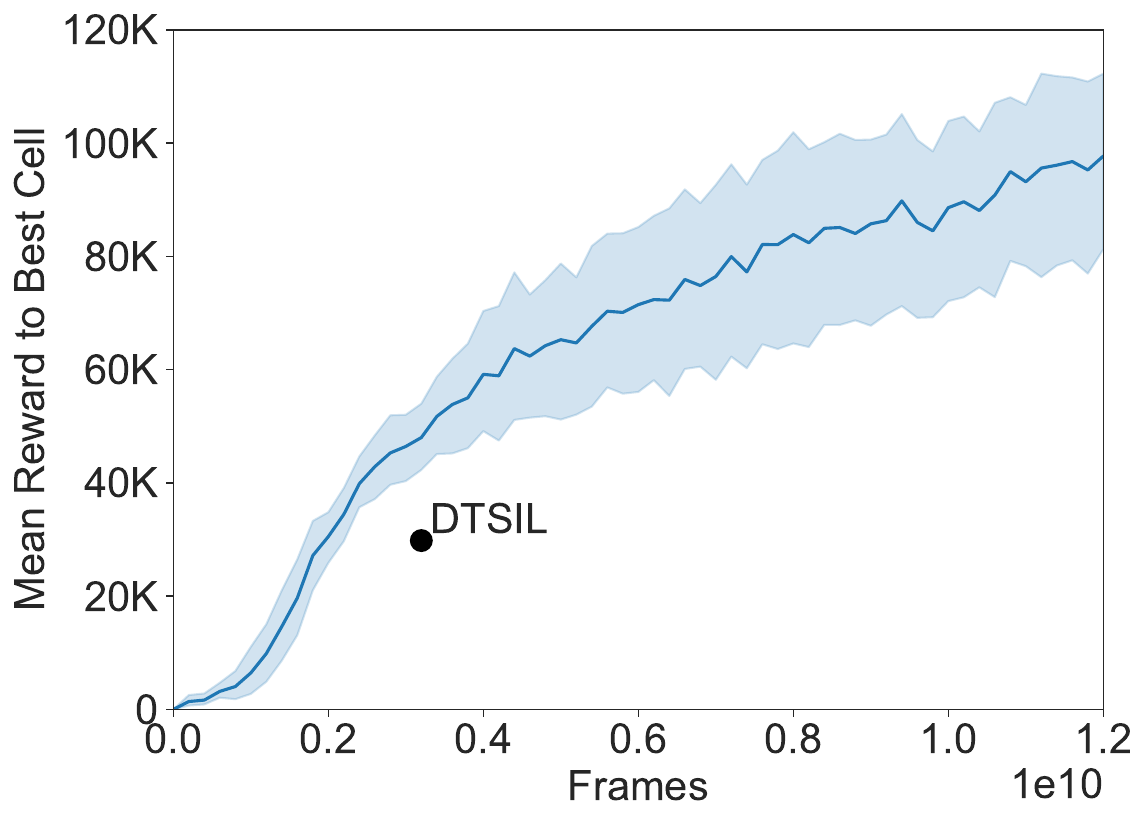}
        \caption{Montezuma's Revenge}
    \end{subfigure}%
    \begin{subfigure}[t]{0.45\textwidth}
        \centering
        \includegraphics[width=\linewidth]{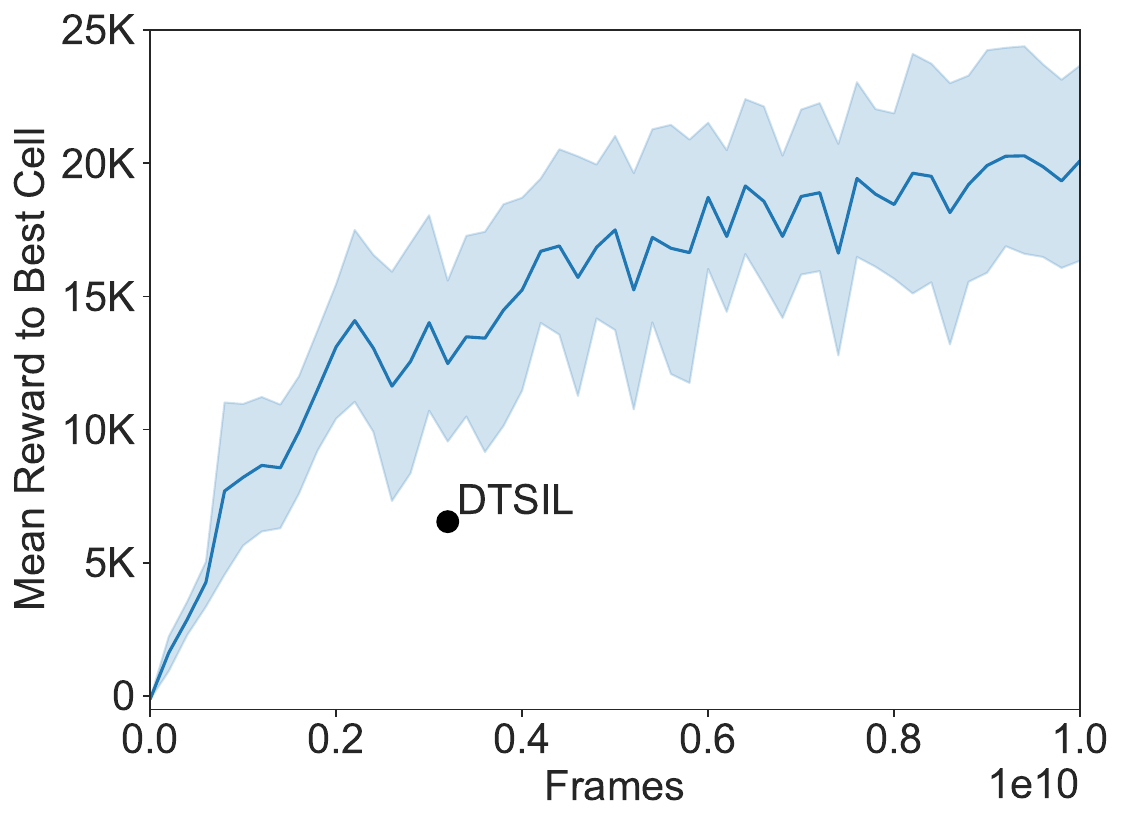}
        \caption{Pitfall}
    \end{subfigure}%
    \caption{\textbf{Policy-based Go-Explore test performance over time compared against final performance of DTSIL.} The final performance of DTSIL is indicated by the black dot and is positioned at 3.2 billion frames, the number of frames for which the DTSIL agent was trained. \textbf{(a)} On Montezuma's Revenge policy-based Go-Explore outperforms DTSIL after 3.2 billion frames by roughly 18,000 points. \textbf{(b)} On Pitfall policy-based Go-Explore outperforms DTSIL after 3.2 billion frames by roughly 6,000 points.}
    \label{fig:policy_ge_dtsil}
\end{figure}

After a pre-print paper describing Go-Explore\cite{ecoffet2019go} (but not policy-based Go-Explore) was published, and after our work on policy-based Go-Explore was long underway, another research team independently developed and published the Diverse Trajectory-conditioned Self-Imitation Learning algorithm (DTSIL)\cite{guo2019efficient}, which is similar to policy-based Go-Explore in many ways, as detailed below. Policy-based Go-Explore outperforms DTSIL on both Montezuma's Revenge and Pitfall after 3.2 billion frames, despite the fact that policy-based Go-Explore was tested on a harder problem, i.e.\ with sticky actions (Fig.~\ref{fig:policy_ge_dtsil}).
In addition, the score of policy-based Go-Explore keeps increasing, eventually achieving a score of 97,728 on Montezuma's Revenge and 20,093 on Pitfall after 12 billion and 10 billion frames, respectively.
It is possible that the performance of DTSIL would also improve with additional frames, but those results were not reported.

DTSIL is similar to policy-based Go-Explore in that it follows the methodology described in the Go-Explore pre-print in the following ways: (1) Like the original Go-Explore, DTSIL explicitly keeps track of an archive of many different states and trajectories to those states, (2) DTSIL first moves the agent to one of these states before performing random exploration, (3)
DTSIL determines whether to add a state to the archive with the help of a domain-knowledge based state embedding, where similar embeddings are grouped into a single cluster (i.e. a cell representation), and (4) DTSIL selects trajectories to follow (i.e. states to return to) by selecting them probabilistically based on the number of times particular clusters have been visited, though DTSIL did not consider any domain knowledge specific information for the purpose of this selection procedure.

Similar to policy-based Go-Explore, and as was recommended as a profitable future direction in our pre-print\cite{ecoffet2019go}, DTSIL is a method that returns to previously visited cells with the help of a goal-conditioned policy (referred to as a trajectory-conditioned policy in the DTSIL paper because the policy is provided with a sequence of the next few goals, as explained below). 
Also similar to policy-based Go-Explore, DTSIL follows a trajectory of intermediate sub-goals towards a particular goal cell, rather than conditioning the policy directly on the state to return to, and DTSIL includes self-imitation learning to make training the policy more sample efficient.
While working on policy-based Go-Explore, we independently invented the technique of following a trajectory of sub-goals and harnessing self-imitation learning.

One major difference between policy-based Go-Explore and DTSIL is that, when following a trajectory, DTSIL aims to provide the entire trajectory as input to the policy, rather than just the next cell in the trajectory.
As a result, the DTSIL network architecture requires some method to deal with variable length trajectories (resolved with an attention layer), while the policy-based Go-Explore architecture only requires the next (sub) goal cell as an input.
Note that, in the DTSIL paper, most trajectories that are followed were found to be too long to provide to the network in their entirety, meaning the trajectory is instead provided to the network in small chunks, resulting in a dynamic that is similar to providing only the next goal.

Another major difference is that policy-based Go-Explore also samples from the policy while exploring, rather than just relying on random actions.
As mentioned before (Extended Data Fig.~\ref{efig:policy_based_ge_cell_discovery}), sampling from the policy results in the discovery of many more cells, potentially explaining the large performance advantage Go-Explore exhibits vs. DTSIL.

A third major difference is the way in which DTSIL and policy-based Go-Explore transition from exploration to exploitation. 
DTSIL transitions from exploration to exploitation by either slowly annealing across training iterations from selecting promising cells for exploration to selecting the highest scoring cells, or by making such a transition abruptly once a particular score threshold is reached.
Policy-based Go-Explore, on the other hand, only focuses on exploration during training. 
Interestingly, despite being asked to return to all cells, rather than spending a good number of training iterations on just the highest performing ones (which is what DTSIL does),
we found that policy-based Go-Explore tends to be able to reliably return to the highest scoring cell in the archive at test time.

A fourth major difference is our introduction of increasing entropy when the agent takes too long to reach the next cell (see Methods).
This entropy increases the exploration performed by policy-based Go-Explore only when necessary, thus largely avoiding the problem of derailment.
Because derailment can severely lower performance, we expect that this innovation contributes substantially to the performance advantage of our implementation of policy-based Go-Explore relative to DTSIL.

A last major difference is with respect to the experiments that were performed. 
DTSIL was tested on Montezuma's Revenge and Pitfall without sticky actions.
In preliminary experiments with policy-based Go-Explore, we found that removing sticky actions greatly simplified the problem, and testing policy-based Go-Explore without sticky actions would have increased its performance.
As a result, and as explained in methods, we did not include DTSIL in our comparison with the state of the art, but we did provide a comparison at the beginning of this section.

Besides these major differences, there are many smaller differences between the two algorithms, including differences in cell selection probabilities, SIL equations, reward clipping, maximum episode length, and hyperparameters.
For a full overview of these differences, we recommend comparing the methods explained in this paper directly with the methods described in the DTSIL paper\cite{guo2019efficient}.

Because the algorithm described in the DTSIL paper is similar to policy-based Go-Explore, in preliminary experiments, we tested whether some of the hyperparameters described in the DTSIL paper would improve the performance of policy-based Go-Explore.
In these preliminary experiments, we found that the grid size of their cell representation (determining the granularity for the x and y coordinates of the agent) of 9x9\footnote{Note that, while an Atari frame is 160x210 pixels, the top 50 rows of pixels are unreachable in our test games and were ignored, meaning that a discretization of 18x18 pixels does result in a 9x9 grid.} and their learning rate of $2.5 \cdot 10^{-4}$ did indeed perform better than the hyperparameters we were testing at that time, and we adopted these hyperparameters from the DTSIL paper instead.

The training performance of DTSIL on Montezuma's Revenge and Pitfall was reported by \citefull{guo2019efficient} for experiments that ran for 3.2 billion frames.
However, the performance-over-time graph presented in the DTSIL paper is not directly comparable with the performance-over-time graph shown in this paper because they represent results for different selection strategies.
For policy-based Go-Explore, we always report the average score achieved when returning to the highest scoring cell in the archive (obtained after training is completed by loading stored checkpoints and testing the policy 100 times).
In contrast, the DTSIL graph shows a rolling average score during training, which means that its average includes returning to low-scoring cells as the algorithm attempts to explore the environment.
That said, the final performance of DTSIL after 3.2 billion frames is measured over only the highest scoring trajectories, and can thus be reasonably compared with the testing performance of policy-based Go-Explore after that many frames.
For Montezuma's Revenge, we compare policy-based Go-Explore against the results that were reported in the supplementary information of the DTSIL paper for a version of DTSIL that implemented the same cell representation as the one used in the policy-based Go-Explore experiments (note, the Montezuma's Revenge results reported in the main DTSIL paper were lower).
For Pitfall, we compare against the only reported results, which were obtained with a slightly different cell representation than the one used by policy-based Go-Explore.
Specifically, the DTSIL cell representation includes the cumulative positive reward achieved; the representations are otherwise the same.

\subsection{No-ops and sticky actions}

The Atari benchmark has been accepted as a common RL benchmark because of the large variety of independent environments it provides \cite{bellemare2013arcade}.
One downside of the games available in this benchmark is that they are inherently deterministic, making it possible to achieve high scores by simply memorizing state-action mappings or a fixed sequence of actions, rather than learning a policy able to generalize to the much larger number of states that are available in each game.
As the community is interested in learning general policies rather than brittle solutions, many have suggested approaches to improve the benchmark, usually by adding some form of stochasticity to the game\cite{Machado2018RevisitingTA}.

One of the first of these approaches was to start each game with a random number (up to 30) of \emph{no-op} (i.e. do nothing) actions\cite{mnih:nature15}.
Executing a random number of no-ops causes the game to start in a slightly different state each episode, as many game entities like enemies and items move in response to time.
While no-ops add some stochasticity at the start of an episode, the game dynamics themselves are still deterministic, allowing for frame-perfect strategies that would be impossible for a human player to reproduce reliably.
In addition, with only 30 different possible starting states, memorization is more difficult, but still possible.

Because of the downsides of no-ops, an alternative approach called \emph{sticky actions} was recommended by the community\cite{Machado2018RevisitingTA}.
Sticky actions mean that, at any time-step greater than 0, there exists a $25\%$ chance that the current action of the agent is ignored, and the previous action is executed instead.
In a way, sticky actions simulate the fact that it is difficult for a human player to provide the desired input at the exact right frame; often a button is pressed a little bit too early, a little bit too late, or held down for a little too long.
Given that Atari games have been designed for human play, this means that human competitive scores should be achievable despite the stochasticity introduced by sticky actions.

While sticky actions have been recommended by the community, no-ops are still widely employed in many recent papers\cite{puigdomenech2020never, badia2020agent57}.
As it is possible that no-ops add some challenges not encountered with just sticky actions, we ensure that Go-Explore is evaluated under conditions that are at least as difficult as those presented in recent papers by evaluating Go-Explore with \emph{both} sticky actions and no-ops. All Go-Explore scores in this paper come from evaluations with both of these forms of stochasticity combined.

\subsection{PPO and SIL}

Both the robustification ``backward'' algorithm and the implementation of policy-based Go-Explore are based on the actor-critic-style PPO algorithm from \citefull{Schulman2017ProximalPO}, wherein $N$ parallel actors collect data in mini-batches of $T$ timesteps, and policy updates are performed after each batch. 
In all PPO-based algorithms presented here, the loss of the policy and value function (both parameterized by $\theta$) is defined as
\begin{equation}
\mathcal{L}(\theta) = \mathcal{L}^{PG}(\theta) + w_{VF} \mathcal{L}^{VF}(\theta) + w_{ENT} \mathcal{L}^{ENT}(\theta) + w_{L2} \mathcal{L}^{L2} + w_{SIL} \mathcal{L}^{SIL}(\theta)
\end{equation}
$\mathcal{L}^{PG}(\theta)$ is the policy gradient loss with PPO clipping, defined as
\begin{equation}
\mathcal{L}^{PG}(\theta) = \mathbb{E}_{s, a \sim \pi_\theta} [\mathrm{max}(-A_t r^\pi_t(\theta), -A_t \mathrm{clip}(r^\pi_t(\theta), 1-\epsilon, 1+\epsilon))]
\end{equation}
\begin{equation}
r^\pi_t(\theta) = \frac{\pi_\theta(a_t | s_t)}{\pi_{\theta_{old}}(a_t | s_t)}
\end{equation}
where $s$ izs a state, $a$ is an action sampled from the policy $\pi_\theta$, $r_t$ is the reward obtained at time-step $t$, and $\epsilon$ is a hyperparameter limiting how much the policy can be changed at each epoch.
$A_t$ is a truncated version of the generalised advantage estimation:
\begin{equation}
    \hat{A}_t = \delta_t + (\gamma \lambda) \delta_{t+1} + ... + (\gamma \lambda)^{T-t+1} \delta_{T-1}
\end{equation}
\begin{equation}
    \delta_{t} = r_t + \gamma V_\theta(s_{t+1}) - V_\theta(s_t)
\end{equation}
where $\gamma$ is the discount factor, $\lambda$ interpolates between a 1-step return and a $T$-step return, $V_\theta$ is the value function parameterised by $\theta$, and $s_t$ and $r_t$ are the state and reward at time step $t$, respectively. 
Similar to the policy gradient loss, the value function loss $\mathcal{L}^{VF}(\theta)$ implemented here also includes PPO-based clipping, and is defined as
\begin{equation}
\mathcal{L}^{VF}(\theta) = \mathbb{E}_{s, a \sim \pi_\theta} [\mathrm{max}((V_\theta(s_t) - \hat{R}_t)^2, (\mathrm{clip}(V_\theta(s_t) - V_{\theta_{old}}(s_t), -\epsilon, \epsilon) - \hat{A}_t)^2)]
\end{equation}
\begin{equation}
\hat{R}_t = V_{\theta_{old}}(s_t, g_t)+\hat{A}_t
\end{equation}
The algorithms further include entropy regularization\cite{o2016combining} $\mathcal{L}^{ENT}(\theta)$ and an L2 regularization penalty $\mathcal{L}^{L2}(\theta)$.

Finally, both the robustification algorithm and policy-based Go-Explore include a self-imitation learning (SIL) loss\cite{Oh2018SelfImitationL}, $\mathcal{L}^{SIL}(\theta)$, which is calculated over previously collected data $\mathcal{D}$. 
While the source of $\mathcal{D}$ differs between the robustification algorithm and policy-based Go-Explore (see their sub-sections for details), in both cases $\mathcal{D}$ comes from previously collected trajectories $\tau$ and consists of tuples $(s, a, R)$, where $s$ is a state encountered in one of these rollouts, $a$ is the action that was taken in that state, and $R$ is the discounted reward that was collected from that state.
With SIL, a small number ($N_{SIL}$) of PPO's actors are assigned to be SIL actors.
Each of these SIL actors, instead of taking actions in the environment, replays one of these trajectories $\tau$, and at each iteration the data collected by these SIL actors, forms the data set $\mathcal{D}$.
The $\mathcal{L}^{SIL}(\theta)$ loss is then calculated as
\begin{equation}
\mathcal{L}^{SIL}(\theta) = \mathcal{L}^{SIL\_PG}(\theta) + w_{SIL\_VF} \mathcal{L}^{SIL\_VF}(\theta) + w_{SIL\_ENT} \mathcal{L}^{SIL\_ENT}(\theta)
\end{equation}
\begin{equation}
\mathcal{L}^{SIL\_PG}(\theta) = \mathbb{E}_{s, a, R \in \mathcal{D}}[-\mathrm{log} \pi_\theta(a | s) \cdot \mathrm{max}(0, R - V_{\theta_{old}}(s))]
\end{equation}
\begin{equation}
\mathcal{L}^{SIL\_VF}(\theta) = \mathbb{E}_{s, a, r \in \mathcal{D}}\left[\frac{1}{2}\mathrm{max}(0, R - V_{\theta}(s))^2\right]
\end{equation}

\noindent
Here, $\mathcal{L}^{SIL\_ENT}(\theta)$ is the entropy regularization term\cite{o2016combining} calculated by evaluating the current policy over $\mathcal{D}$.

All architectures that feature recurrent units are updated with a variant of the truncated back-propagation-through-time algorithm called BPTT$(h; h')$\cite{williams1990efficient}, which is suitable when data is gathered in mini-batches. In BPTT$(h; h')$, the network is updated every $h'$ timesteps (here $h'=T$, the number of timesteps in the mini-batch), but it is unfolded for $h \geq h'$ timesteps (here $h=1.5T$), meaning that gradients can flow beyond the boundaries of the mini-batch. To facilitate BPTT$(h; h')$, the first mini-batch of each run consists of $1.5T$ timesteps.

\subsection{The backward algorithm}
\label{sec:backward}

The ``backward algorithm''\cite{salimans2018learning} operates by placing the agent close to the end of the trajectory and running PPO until the performance of the agent matches or exceeds that of the demonstration.
Once that is achieved, the agent's starting point is moved closer to the trajectory's beginning and the process is repeated.
The algorithm thus learns the demonstration in a backward order.

The original version of the backward algorithm relied on a single demonstration due to the assumption that obtaining human demonstrations was expensive.
In our case, however, obtaining multiple demonstrations is easy and cheap by simply re-running the exploration phase, which is why we modified the algorithm to utilise multiple demonstrations: at the start of each episode, a demonstration is chosen at random to provide the starting point for the agent.
For Atari, 10 demonstrations from different runs of the exploration phase were used for each robustification. 
In Atari, the demonstration was extracted by finding all trajectories that reached an end-of-episode state (to prevent selection of length 0 trajectories in games with exclusively negative rewards, see Methods ``\nameref{sec:evaluation}''), and extracting the shortest one among those with the highest score.
For robotics, because it was possible to extract diverse demonstrations from a single run of the exploration phase (this was not possible in Atari because high-scoring trajectories within a given Atari run tended to share most of their actions), 10 demonstrations from the same runs were used for robustification. In robotics, the first demonstration corresponds to the shortest successful trajectory (i.e.~the shortest trajectory that puts the object in the shelf), while each subsequent demonstration corresponds to the successful trajectory with the highest mean difference from all previously selected trajectories, where the difference between two trajectories is given by $\frac{\sum_{i=1}^{L}I(\tau^a_i \ne \tau^b_i)}{L}$ ($\tau^a$ and $\tau^b$ are the list of actions being compared, $L = \min(|\tau^a|, |\tau^b|)$, $I$ is the indicator function). 
Because actions are continuous, meaning that it is exceedingly unlikely for two independently sampled actions to be the same, $\tau^a_i = \tau^b_i$ only for parts where the trajectories are identical because they were branched from the same intermediate trajectory.
As such, this metric effectively measures to what degree the two trajectories have a shared history, and prefers trajectories that share as little history as possible.
In both cases, SIL was also performed on the set of demonstrations provided to the backward algorithm (see ``PPO and SIL'' section).
Each robustification run used demonstrations from different, non-overlapping exploration phase runs (10 exploration runs for Atari, 1 for robotics).

On Atari, the score that can be obtained by starting the algorithm from the start of the environment is tracked throughout the run by adding a virtual demonstration of length 0, i.e.\ traditional training that executes the current policy from the domain's traditional starting state. This addition makes it possible to occasionally obtain superhuman policies even when the backward algorithm has not yet reached the starting point of any of the non-virtual demonstrations it was provided. During training, the time limit of an episode is the remaining length of the demonstration (which is generally much shorter than the environment time limit, especially at the beginning of training since we start at the end of demonstrations and move backwards)        plus a few extra frames (Extended Data Table~\ref{etab:backward_hyper_atari}). When the virtual demonstration is selected, however, the time limit is that of the underlying environment (Extended Data Table~\ref{etab:atari_hyper}). As a result, training episodes in which the virtual demonstration was selected often require many more frames to complete than those corresponding to an exploration phase demonstration.  To balance the number of frames allocated to the virtual demonstration, the average number of steps in an episode corresponding to the virtual demonstration ($l_v$) is tracked as well as the average number of steps corresponding to starting from any other demonstration ($l_d$), and the selection probability of the virtual demonstration is then $\frac{1}{11} \frac{l_d}{l_v}$, where 11 is the total number of demonstrations (10 from the exploration phase runs and 1 virtual demonstration). In cases where the virtual demonstration was not stochastically chosen, one of the exploration phase demonstrations was chosen uniformly at random.

A key difficulty in implementing an RL algorithm that can perform well across all Atari games with identical hyperparameters is the significant variations in reward scales within the Atari benchmarks, with some games having an average reward of 1 and others with average rewards of over 1,000. 
Traditionally, this challenge has been addressed with reward \emph{clipping}, in which all rewards are clipped to either -1 or +1, but such an approach is problematic in games (e.g.\ Pitfall and Skiing) in which the scale of rewards within the game is relevant because it tells the agent the relative importance of different rewarded (or punished) actions. 
In this work, we take advantage of the fact that the deterministic exploration phase is unaffected by reward scale and can provide us with a sense of the scale of scores achievable in each game. 
We are thus able to use reward \emph{scaling} in the robustification phase: at the start of the robustification phase, the rewards of the demonstrations are used to produce a reward multiplier that will result every game having approximately the same value function scale.
This reward multiplier is given by
\begin{equation}
    m = \frac{C}{\mu_V}
\end{equation}
where $C$ is a constant representing the target average absolute value of the value function when following the demonstration (in all our experiments, $C = 10$), and $\mu_V$ is defined as
\begin{equation}
    \mu_V = \frac{1}{\sum_{d=1}^D T_d}\sum_{d=1}^D \sum_{t=1}^{T_d} |V_d(t)|
\end{equation}
where $D$ is the number of demonstrations, $T_d$ is the number of steps in each demonstration, and $V_d(t)$ is the sum of discounted rewards in demonstration $d$ starting from step $t$.%

\subsection{Go-Explore vs. Agent57}

The creation of the Atari benchmark started the search for reinforcement learning algorithms capable of achieving super-human performance on all games in this benchmark\cite{bellemare2013arcade}.
For a majority of these games, super-human performance was reached quickly through early deep reinforcement learning techniques now considered standard\cite{mnih:nature15}, but for a small set of games super-human performance remained out of reach.
The work presented in this paper achieved this historic feat concurrently with an algorithm called Agent57\cite{badia2020agent57}.
Go-Explore and Agent57 accomplish this milestone via very different methods, offering the scientific community a diversity of promising tools to use and build upon going forward.

Agent57 is built upon the Never Give Up (NGU) algorithm\cite{puigdomenech2020never}.
NGU was able to achieve superhuman performance on the majority of the Atari games by combining within-episode and across-episode intrinsic motivation, tracking many Q-functions that each maintain a different trade-off between intrinsic and extrinsic motivation, and implementing efficient parallelization of data collection. 
Agent57 elevated the performance of NGU to superhuman on all games by dynamically learning which of its many Q-functions provides the highest cumulative reward, stabilizing the learning of those Q-functions, and by running the algorithm for an impressive 100 billion frames.
So, while Agent57 achieved the milestone of superhuman performance on the last few remaining games at the same time as Go-Explore, its method is vastly different from Go-Explore.

With respect to results, it is first of all important to reiterate that Go-Explore was evaluated in an environment with sticky actions (i.e. following community standards, see SI ``No-ops and sticky actions'') while Agent57 was evaluated in an environment without sticky actions.
Sticky actions make the games substantially harder to play well, which is why Agent57 was not considered for direct comparison with Go-Explore in the main paper (see Methods ``State of the art on Atari'').

Despite the fact that Go-Explore solutions were evaluated under more difficult conditions, Go-Explore still outperforms Agent57 on 7 out of the 11 games that we tested (Table~\ref{tab:ge_vs_agent57}).
It is also worth noting that the Go-Explore results reported here were obtained after a total of 30 billion (or 40 billion for Solaris) frames of training data, while Agent57 was trained for 100 billion frames.
While Go-Explore does ``skip'' frames by reloading simulator state, we argue that these frames would also be skipped in almost any scenario where Go-Explore is practically applied, as it should be possible to save and restore the state of a modern simulator.
That said, the relative sample efficiency of policy-based Go-Explore (e.g. 97,728 points on Montezuma's Revenge after 12 billion frames) suggests that policy-based Go-Explore could be more sample efficient than Agent57 even if states can not be restored, though the fact that policy-based Go-Explore was only tested with domain knowledge makes it impossible to provide a fair comparison at this time.

\begin{table}
\centering
\fontsize{9}{10.5}\selectfont
\begin{tabular}{l|rr}{} 
Game &     Go-Explore & Agent57  \\
\hline
Berzerk          &   \textbf{197,376} &       61,508 \\
Bowling          &       \textbf{260} &          251 \\
Centipede        & \textbf{1,422,628} &      412,848 \\
Freeway          &        \textbf{34} &           33 \\
Gravitar         &     7,588 &       \textbf{19,214} \\
MontezumaRevenge &    \textbf{43,791} &        9,352 \\
Pitfall          &     6,954 &       \textbf{18,756} \\
PrivateEye       &    \textbf{95,756} &       79,717 \\
Skiing           &    \textbf{-3,660} &       -4,203 \\
Solaris          &    19,671 &       \textbf{44,200} \\
Venture          &     2,281 &        \textbf{2,628} \\
\end{tabular}
\caption{\textbf{Go-Explore outperforms Agent57 on 7 out of the 11 games that we tested.} Here we show the results of the Go-Explore variant where the exploration phase was performed without domain knowledge and with restoration of simulator state. The Go-Explore results were obtained by re-evaluating the final agent 1,000 times on the environment with sticky actions and no-ops. }
\label{tab:ge_vs_agent57}
\end{table}

\subsection{ALE issues}
\label{sec:ale_issues}

While the Arcade Learning Environment (ALE)\cite{bellemare2013arcade}, which is the underlying backend of OpenAI Gym, is the the standard way to interface with Atari games in RL, the library comes with a couple issues that needed to be addressed in our work.

First, the score on Montezuma's Revenge rolls over (i.e.\ is subject to numerical overflow) when it exceeds 1 million, which is incorrectly interpreted by the ALE as a negative reward of -1 million.
We patched the environment to remove this bug and thereby make it possible for algorithms to learn to produce scores higher than 1 million. This enabled us to learn that Go-Explore can substantially outperform the human world record of 1.2 million\cite{atari_scoreboard}.
No previous work had achieved scores anywhere near high enough to trigger this particular bug on Montezuma's Revenge.

Second, the implementation of Montezuma's Revenge in the ALE library includes a bug that prevents the agent from progressing to the next level when the agent is on its last life, which is clearly unintended behaviour that does not occur in the original game.
Because there are no penalties for losing a life, policy-based Go-Explore learns to sacrifice lives in order to bypass hazards or to return to the entrance of a room more quickly.
As a result, policy-based Go-Explore frequently reaches the treasure room without any lives remaining, preventing further progress.
As such, for policy-based Go-Explore only, we terminate the episode on first death, which avoids this bug without simplifying the game.

\subsection{Infrastructure}

In terms of infrastructure, each exploration phase run was performed on a single worker machine equipped with 44 CPUs and 96GB of RAM, though memory usage is substantially lower for most games. Each robustification run was parallelized across 8 worker machines each equipped with 11 CPUs, 24GB of RAM, and 1 GPU. 
Policy-based Go-Explore was parallelized across 16 worker machines each equipped with 11 CPUs, 10GB of RAM, and 1 GPU.

\FloatBarrier

\printbibliography[segment=3,check=onlynew]

\end{document}